\documentclass[11pt,draftclsnofoot,onecolumn]{IEEEtran}

\usepackage{graphicx}
\usepackage{epsfig}
\usepackage{amsmath}
\usepackage{amssymb}
\usepackage{subfigure}
\usepackage{wrapfig}
\usepackage{times,color}
\usepackage{cite,footnote,xspace,syntonly,algorithm,algorithmic,bm}
\input{mysymbol.sty}
\long\def\symbolfootnote[#1]#2{\begingroup
\def\thefootnote{\fnsymbol{footnote}}
\footnote[#1]{#2}\endgroup} \psfull

\begin{document}
%\setlength{\abovecaptionskip}{-7pt}

% % % % % % % % % % % % % % % % % % % % % % % % % % % % % % % % % % % % % % % %
%                         Cover Page                                          %
% % % % % % % % % % % % % % % % % % % % % % % % % % % % % % % % % % % % % % % %

\title{\huge Robust PCA as Bilinear Decomposition\\
 with Outlier-Sparsity Regularization$^\dag$}

\author{{\it Gonzalo~Mateos and Georgios~B.~Giannakis~(contact author)$^\ast$}}

\markboth{IEEE TRANSACTIONS ON SIGNAL PROCESSING (SUBMITTED)}
\maketitle \maketitle \symbolfootnote[0]{$\dag$ This work was supported by 
MURI (AFOSR FA9550-10-1-0567) grant. Part of the paper appeared in the 
{\it Proc. of
the 44th Asilomar Conference on Signals, Systems, and
Computers}, Pacific Grove, CA, Nov. 7-10, 2010.}
\symbolfootnote[0]{$\ast$ The authors are with the Dept. of
Electrical and Computer Engineering, University of Minnesota,
200 Union Street SE, Minneapolis, MN 55455. Tel/fax:
(612)626-7781/625-2002; Emails:
\texttt{\{mate0058,georgios\}@umn.edu}}

\vspace*{-80pt}
\begin{center}
\small{\bf Submitted: }\today\\
\end{center}
\vspace*{10pt}

\thispagestyle{empty}\addtocounter{page}{-1}

% % % % % % % % % % % % % % % % % % % % % % % % % % % % % % % % % % % % % % % %
%                         Abstract                                            %
% % % % % % % % % % % % % % % % % % % % % % % % % % % % % % % % % % % % % % % %

\begin{abstract}
Principal component analysis (PCA) is widely used for
dimensionality reduction, with well-documented
merits in various applications involving high-dimensional
data, including computer vision, preference measurement, and
bioinformatics. In this context, the fresh look advocated here
permeates benefits from variable selection and compressive
sampling, to robustify PCA against outliers. A least-trimmed
squares estimator of a low-rank bilinear factor analysis model is
shown closely related to that obtained from an
$\ell_0$-(pseudo)norm-regularized criterion encouraging
\textit{sparsity} in a matrix explicitly modeling the outliers.
This connection suggests robust PCA schemes based
on convex relaxation, which lead naturally to a family of
robust estimators encompassing Huber's optimal M-class as a special case. 
Outliers are identified by tuning a regularization parameter, which
amounts to controlling sparsity of the outlier matrix along
the whole \textit{robustification} path of (group) least-absolute shrinkage
and selection operator (Lasso)
solutions. Beyond its neat ties to robust statistics, the developed 
outlier-aware PCA framework is versatile 
 to accommodate novel and scalable algorithms to: i) 
track the low-rank signal subspace robustly, as new data are acquired in real 
time; and ii) determine principal components robustly in (possibly)
infinite-dimensional feature spaces. Synthetic and real 
data tests corroborate the effectiveness of the proposed robust PCA schemes, 
when 
used to identify aberrant responses in personality assessment surveys, as well 
as 
unveil communities in social networks, and intruders from video surveillance 
data.
\end{abstract}
\vspace*{-5pt}
\begin{keywords}
Robust statistics, principal component analysis, outlier rejection,
sparsity, (group) Lasso.
\end{keywords}
%\newpage
\begin{center} \bfseries EDICS Category: MLR-LEAR\end{center}
%
% for peerreview papers, inserts a page break and creates the second title.
% Will be ignored for other modes.
%\IEEEpeerreviewmaketitle
\newpage

% % % % % % % % % % % % % % % % % % % % % % % % % % % % % % % % % % % % % % % %
%                         Section I                                           %
% % % % % % % % % % % % % % % % % % % % % % % % % % % % % % % % % % % % % % % %

\section{Introduction}\label{sec:intro}

Principal component analysis (PCA) is the workhorse of
high-dimensional data analysis and dimensionality reduction,
with numerous applications in statistics, engineering, and the
biobehavioral sciences; see, e.g.,~\cite{J02}. Nowadays ubiquitous
e-commerce sites, the Web, and urban traffic
surveillance systems generate massive volumes of data. As a
result, the problem of extracting the most informative, yet
low-dimensional structure from high-dimensional datasets is
of paramount importance~\cite{HTF09}. To this end,  PCA
provides least-squares (LS) optimal linear approximants in $\mathbb{R}^q$
to a data set in $\mathbb{R}^p$, for $q\leq p$. The desired
linear subspace is obtained from the $q$ dominant
eigenvectors of the sample data covariance matrix~\cite{J02}.

Data obeying postulated low-rank models include also outliers, which are 
samples not adhering to those nominal models. Unfortunately, LS is known to be 
very sensitive to outliers~\cite{RL87,hr09}, and this undesirable
property is inherited by PCA as well~\cite{J02}. Early efforts
to robustify PCA have relied on robust estimates of the
data covariance matrix; see, e.g.,~\cite{c80}. 
Related approaches are driven from statistical physics~\cite{xu95}, and also 
from M-estimators~\cite{dlt03}. Recently, polynomial-time algorithms
with remarkable performance guarantees have emerged for low-rank matrix
recovery in the presence of sparse -- but otherwise arbitrarily large --
 errors~\cite{clmw09,cspw}. This
pertains to an `idealized robust' PCA setup, since those
entries not affected by outliers are assumed error free.  Stability in
reconstructing the low-rank and sparse matrix components
in the presence of `dense' noise have been reported 
in~\cite{zlwcm10,Outlier_pursuit}. 
A hierarchical Bayesian model was proposed to tackle the aforementioned 
low-rank plus sparse matrix decomposition problem in~\cite{bayes_rpca}.

In the present paper, a robust PCA approach is pursued
requiring minimal assumptions on the outlier model. A natural
least-trimmed squares (LTS) PCA estimator is first shown
closely related to an estimator obtained from an
$\ell_0$-(pseudo)norm-regularized criterion, adopted to fit a
low-rank bilinear factor analysis model that explicitly incorporates
an unknown \textit{sparse} vector of outliers per datum (Section 
\ref{sec:robust_problem}). As in
compressive sampling~\cite{tropp06tit}, efficient (approximate)
solvers are obtained in Section \ref{sec:spacor}, by surrogating 
the $\ell_0$-norm
of the outlier matrix with its closest convex approximant. 
This leads naturally to an M-type PCA estimator, which subsumes
Huber's optimal choice as a special case~\cite{fuchs99}. Unlike
Huber's formulation though, results here are not confined to an
outlier contamination model. A tunable parameter controls the
sparsity of the estimated matrix, and the number of outliers as
a byproduct. Hence, effective data-driven methods to select this parameter
are of paramount importance, and systematic approaches are
pursued by efficiently exploring the entire
\textit{robustifaction} (a.k.a. homotopy) path of (group-) Lasso
solutions~\cite{HTF09,yl06grouplasso}. In this sense, the
method here capitalizes on but \textit{is not limited to}
sparse settings where outliers are sporadic, since one can
examine all sparsity levels along the robustification path. The outlier-aware 
generative data model and its sparsity-controlling estimator are quite 
general, since minor modifications discussed in Section \ref{ssec:connections} 
enable robustifiying linear regression~\cite{uspacor}, dictionary 
learning~\cite{Dictionary_learning_SP_mag_10,mairal10}, and K-means clustering 
as well~\cite{HTF09,pf_vk_gg_clustering}. Section 
\ref{sec:algo_issues} deals with further 
modifications for bias reduction through nonconvex 
regularization, and automatic determination of the reduced dimension 
$q$.

Beyond its neat ties to robust statistics, the developed 
outlier-aware PCA framework is 
versatile to accommodate scalable \textit{robust} algorithms to: i) 
track the low-rank signal subspace, as new data are 
acquired in real time (Section \ref{sec:online}); 
and ii) determine principal components in (possibly)
infinite-dimensional feature spaces, thus robustifying kernel 
PCA~\cite{ssm97}, 
and spectral clustering as well~\cite[p. 544]{HTF09} (Section 
\ref{sec:kernel_PCA}). The vast 
literature on \textit{non-robust} subspace 
tracking algorithms includes~\cite{yang95,mairal10}, 
and~\cite{balzano_tracking}; see also~\cite{GRASTA} for a first-order 
algorithm that is robust to outliers and incomplete data. 
Relative to~\cite{GRASTA}, the online robust (OR-) PCA algorithm of this 
paper is a second-order method, which minimizes an outlier-aware 
exponentially-weighted LS estimator of the low-rank factor analysis model.  
Since the outlier 
and subspace estimation tasks decouple nicely in OR-PCA, one can readily 
devise 
a first-order counterpart when minimal computational loads are at a 
premium. In terms of performance, online algorithms 
are known to be markedly faster than their batch 
alternatives~\cite{balzano_tracking,GRASTA}, e.g., in the timely 
context of low-rank matrix 
completion~\cite{Recht_SIAM_2010,Recht_Parallel_2011}. While 
the focus here is not on incomplete data records, extensions to account for 
missing data are immediate and will be reported elsewhere. 

In Section \ref{sec:sims}, numerical 
tests with synthetic and real 
data corroborate the effectiveness of the proposed robust PCA schemes, 
when used to identify aberrant responses from a questionnaire designed to 
measure the Big-Five dimensions of personality traits~\cite{BF_chapter},  
as well as unveil communities in a (social) network of college football 
teams~\cite{network_data}, and intruders from video 
surveillance data~\cite{dlt03}. Concluding remarks are given in Section 
\ref{sec:conclusion}, while a few technical details are deferred to the 
Appendix.

% % % % % % % % % % % % % % % % % % % % % % % % % % % % % % % % % % % % % % % %
%                               Notation                                      %
% % % % % % % % % % % % % % % % % % % % % % % % % % % % % % % % % % % % % % % %

\noindent\textit{Notation:} Bold uppercase (lowercase) letters will denote
matrices (column vectors). Operators $(\cdot)^{\prime}$, $\mbox{tr}(\cdot)$, 
$\textrm{med}(\cdot)$, and $\odot$ will denote transposition, matrix trace,
median, and Hadamard product,
respectively. Vector $\textrm{diag}(\mathbf{M})$ collects the
diagonal entries of $\mathbf{M}$, whereas the diagonal matrix
$\textrm{diag}(\mathbf{v})$ has the entries of $\mathbf{v}$
on its diagonal. The $\ell_p$-norm of 
$\mathbf{x}\in\mathbb{R}^n$ is
$\|\mathbf{x}\|_p:=\left(\sum_{i=1}^n|x_i|^p\right)^{1/p}$ for
$p\geq 1$; and
$\|\mathbf{M}\|_F:=\sqrt{\mbox{tr}\left(\mathbf{M}\mathbf{M}^{\prime}\right)}$
is the matrix Frobenious norm. The $n\times n$ identity
matrix will be represented by $\mathbf{I}_{n}$, while
$\mathbf{0}_{n}$ will denote the $n\times 1$ vector of all
zeros, and $\mathbf{0}_{n\times
m}:=\mathbf{0}_{n}\mathbf{0}_{m}^{\prime}$. Similar notation will 
be adopted for vectors (matrices) of all ones. The $i$-th vector of the 
canonical basis in $\mathbb{R}^n$ will be denoted by $\bbb_{n,i}$, 
$i=1,\ldots,n$.

% % % % % % % % % % % % % % % % % % % % % % % % % % % % % % % % % % % % % % % %
%                         Section II                                          %
% % % % % % % % % % % % % % % % % % % % % % % % % % % % % % % % % % % % % % % %

%\vspace{-0.25cm}
\section{Robustifying PCA}\label{sec:robust_problem}

% % % % % % % % % % % % % % % % % % % % % % % % % % % % % % % % % % % % % % % %
%                         Subsection II-A                                     %
% % % % % % % % % % % % % % % % % % % % % % % % % % % % % % % % % % % % % % % %

%\subsection{Low-rank factor analysis model approach to 
%PCA}\label{ssec:factor_model}
Consider the standard PCA formulation, in which a set
of data $\mathcal{T}_y:=\{\mathbf{y}_n\}_{n=1}^{N}$ in the
$p$-dimensional Euclidean \textit{input} space is given, and
the goal is to find the best $q$-rank $(q\leq p)$ linear
approximation to the data in $\mathcal{T}_y$; see e.g.,~\cite{J02}. 
Unless otherwise stated, it is 
assumed throughout that the value of $q$ is given. One approach to
solving this problem, is to adopt a low-rank bilinear (factor
analysis) model
\begin{equation}\label{eq:factor_model}
\mathbf{y}_n=\mathbf{m}+\mathbf{U}\mathbf{s}_n+\mathbf{e}_n,\quad n=1,\ldots,N
\end{equation}
where $\mathbf{m}\in\mathbb{R}^p$ is a location (mean) vector;
matrix $\mathbf{U}\in\mathbb{R}^{p\times q}$ has orthonormal
columns spanning the signal subspace;
$\{\mathbf{s}_n\}_{n=1}^N$ are the so-termed \textit{principal
components}, and $\{\mathbf{e}_n\}_{n=1}^N$ are zero-mean
i.i.d. random errors. The unknown variables in \eqref{eq:factor_model}
can be collected in
$\mathcal{V}:=\{\mathbf{m},\mathbf{U},\{\mathbf{s}_n\}_{n=1}^N\}$,
and they are estimated using the LS criterion as
\begin{equation}\label{eq:PCA_LS_cost}
\min_{\mathcal{V}}\sum_{n=1}^N\|\mathbf{y}_n-\mathbf{m}-\mathbf{U}\mathbf{s}_n
\|_2^2,\quad\textrm{s. to }\quad\mathbf{U}^\prime\mathbf{U}=\mathbf{I}_q.
\end{equation}
PCA in \eqref{eq:PCA_LS_cost} is a nonconvex optimization
problem due to the bilinear terms $\bbU\bbs_n$, yet a global
optimum $\hat{\mathcal{V}}$ can be shown to exist; 
see e.g.,~\cite{yang95}. The resulting estimates are
$\hat{\mathbf{m}}=\sum_{n=1}^N\mathbf{y}_n/N$ and
$\hat{\mathbf{s}}_n=\hat{\mathbf{U}}^\prime(\mathbf{y}_n-\hat{\mathbf{m}}),\:n=
1,\ldots,N$;
while $\hat{\mathbf{U}}$ is formed with columns equal to the $q$-dominant right
singular vectors of the $N\times p$ data matrix
$\mathbf{Y}:=[\mathbf{y}_1,\ldots,\mathbf{y}_N]^\prime$~\cite[p.
535]{HTF09}. The principal components (entries of)
$\mathbf{s}_n$ are the projections of the centered data 
points $\{\bby_n-\hat{\bbm}\}_{n=1}^N$ onto the signal subspace. 
Equivalently, PCA can be formulated based on maximum variance, 
or, minimum reconstruction error criteria; see e.g.,~\cite{J02}.

% % % % % % % % % % % % % % % % % % % % % % % % % % % % % % % % % % % % % % % %
%                         Subsection II-B                                     %
% % % % % % % % % % % % % % % % % % % % % % % % % % % % % % % % % % % % % % % %

\subsection{Least-trimmed squares PCA}\label{ssec:LTS_PCA}
Given training data $\mathcal{T}_x:=\{\mathbf{x}_n\}_{n=1}^{N}$ possibly 
contaminated with
outliers, the goal here is to develop a robust estimator of
$\mathcal{V}$ that requires minimal assumptions on the outlier
model. Note that there is an explicit notational differentiation between: i) 
the 
data in $\mathcal{T}_y$ which adhere to the nominal model 
\eqref{eq:factor_model}; and ii) the given data in $\mathcal{T}_x$ that  may 
also
contain outliers, i.e., those 
$\mathbf{x}_n$ not adhering to \eqref{eq:factor_model}. Building on LTS 
regression~\cite{RL87}, the desired
robust estimate
$\hat{\mathcal{V}}_{LTS}:=\{\hat{\mathbf{m}},\hat{\mathbf{U}},\{\hat{\mathbf{s}}_n\}_{n=1}^N\}$
for a prescribed $\nu<N$
can be obtained via the following LTS PCA estimator [cf. \eqref{eq:PCA_LS_cost}]
\begin{equation}\label{eq:LTS}
\hat{\mathcal{V}}_{LTS}:=\arg\min_{\mathcal{V}}\sum_{n=1}^{\nu}r_{[n]}^2(\mathcal{V})
,\quad\textrm{s. to }\quad\mathbf{U}^\prime\mathbf{U}=\mathbf{I}_q
\end{equation}
where $r_{[n]}^2(\mathcal{V})$ is the $n$-th order statistic
among the squared residual norms
$r_{1}^2(\mathcal{V}),\ldots,r_{N}^2(\mathcal{V})$, and
$r_n(\mathcal{V}):=\|\mathbf{x}_n-\mathbf{m}-\mathbf{U}\mathbf{s}_n\|_2$.
The so-termed \textit{coverage} $\nu$ determines the breakdown
point of the LTS PCA estimator~\cite{RL87}, since the $N-\nu$
largest residuals are absent from the estimation criterion in \eqref{eq:LTS}. Beyond
this universal outlier-rejection property, the LTS-based estimation offers
an attractive alternative to robust linear regression due to its high breakdown point and
desirable analytical properties, namely $\sqrt{N}$-consistency and
asymptotic  normality under mild assumptions~\cite{RL87}. 
\begin{remark}[Robust estimation of the mean]\label{rem:mean}
In most applications of PCA, data in $\mathcal{T}_y$ are typically assumed 
zero mean.  
This is without loss of generality, since nonzero-mean training data
can always be rendered zero mean, by subtracting the
sample mean $\sum_{n=1}^N\mathbf{y}_n/N$ from each $\bby_n$. 
In modeling zero-mean data, the known vector $\bbm$ in \eqref{eq:factor_model} can
obviously be neglected. When outliers are present however, data in
$\mathcal{T}_x$ are not necessarily zero mean, and it is unwise to center them 
using the non-robust sample mean estimator which has a breakdown 
point equal to zero~\cite{RL87}. Towards robustifying PCA, a more sensible 
approach
is to estimate $\bbm$ robustly, and jointly with $\bbU$ and the principal components
 $\{\bbs_n\}_{n=1}^N$.
\end{remark}

Because
\eqref{eq:LTS} is a nonconvex optimization problem, a
nontrivial issue pertains to the existence of the proposed LTS PCA
estimator, i.e., whether or not \eqref{eq:LTS}
attains a minimum. Fortunately, the answer is in the affirmative
 as asserted next.
\begin{property}\label{prope:existence_LTS} \emph{The LTS PCA estimator is well defined,
since \eqref{eq:LTS} has (at least) one solution.}
\end{property}
Existence of 
$\hat{\mathcal{V}}_{LTS}$ can be readily established as follows: i) for
each subset of $\mathcal T$ with cardinality $\nu$ (there are
${{N}\choose{\nu}}$ such subsets), solve the corresponding PCA
problem to obtain a unique candidate estimator per subset; and ii)
pick $\hat{\mathcal{V}}_{LTS}$ as the one among all
${{N}\choose{\nu}}$ candidates with the minimum cost.

Albeit conceptually simple, the
solution procedure outlined under Property \ref{prope:existence_LTS} 
is combinatorially complex, and thus
intractable except for small sample sizes $N$. Algorithms to
obtain approximate LTS solutions in large-scale linear regression problems are
available; see e.g.,~\cite{RL87}.

% % % % % % % % % % % % % % % % % % % % % % % % % % % % % % % % % % % % % % % %
%                         Subsection II-C                                     %
% % % % % % % % % % % % % % % % % % % % % % % % % % % % % % % % % % % % % % % %

\subsection{$\ell_0$-norm regularization for robustness}\label{ssec:LTS_equiv}

Instead of discarding large residuals, the alternative approach
here explicitly accounts for outliers in the low-rank data
model \eqref{eq:factor_model}. This becomes possible through the vector 
variables
$\{\mathbf{o}_n\}_{n=1}^{N}$ one per training datum $\bbx_n$, which
take the value $\mathbf{o}_n\neq\mathbf{0}_p$ whenever datum
$n$ is an outlier, and $\mathbf{o}_n =\mathbf{0}_p$ otherwise.
Thus, the novel outlier-aware factor analysis model is
\begin{equation}\label{eq:model_outliers}
\mathbf{x}_n=\bby_n+\bbo_n=\mathbf{m}+\mathbf{U}\mathbf{s}_n+\mathbf{e}_n+\mathbf{o}_n, \quad\quad 
n=1,\ldots, N
\end{equation}
where $\mathbf{o}_n$ can be deterministic or random with
unspecified distribution. In the \textit{under-determined}
linear system of equations \eqref{eq:model_outliers}, both
$\mathcal{V}$ as well as the $N\times p$ matrix
$\mathbf{O}:=[\mathbf{o}_1,\ldots,\mathbf{o}_N]^\prime$ are
unknown. The percentage of outliers dictates the degree of
\emph{sparsity} (number of zero rows) in $\mathbf{O}$.
Sparsity control will prove instrumental in efficiently
estimating $\mathbf{O}$, rejecting outliers as a byproduct, and
consequently arriving at a \textit{robust} estimator of
$\mathcal{V}$. To this end, a natural criterion for controlling outlier
sparsity is to seek the estimator [cf. \eqref{eq:PCA_LS_cost}]
\begin{equation}\label{eq:cost_l0unconstr}
\{\hat{\mathcal{V}},\hat{\mathbf{O}}\}=\arg\min_{\mathcal{V},\mathbf{O}}
\|\mathbf{X}-\mathbf{1}_N\mathbf{m}^\prime-
\mathbf{S}\mathbf{U}^\prime-\mathbf{O}\|_F^2
+\lambda_0\|\mathbf{O}\|_0,\quad\textrm{s. to }\mathbf{U}^\prime\mathbf{U}=\mathbf{I}_q
\end{equation}
where 
$\mathbf{X}:=[\mathbf{x}_1,\ldots,\mathbf{x}_N]^\prime\in\mathbb{R}^{N\times
p}$, 
$\mathbf{S}:=[\mathbf{s}_1,\ldots,\mathbf{s}_N]^\prime\in\mathbb{R}^{N\times
q}$, and $\|\mathbf{O}\|_0$ denotes the nonconvex
$\ell_0$-norm that is equal to the number of nonzero
rows of $\mathbf{O}$. %For $p=1$, $\mathbf{O}$ collapses to a column
%vector $\mathbf{o}$, and $\|\mathbf{o}\|_0$ counts the
%nonzero entries of $\mathbf{o}$.
Vector (group) sparsity in the rows $\hat{\mathbf{o}}_n$ of
$\hat{\mathbf{O}}$ can be directly controlled by tuning the
parameter $\lambda_0\geq 0$.

As with compressive sampling and sparse modeling schemes that
rely on the $\ell_0$-norm~\cite{tropp06tit}, the robust PCA
problem \eqref{eq:cost_l0unconstr} is NP-hard~\cite{Natarajan_NP_duro}. In 
addition, the
sparsity-controlling estimator \eqref{eq:cost_l0unconstr} is
intimately related to LTS PCA, as asserted next.

\begin{proposition}\label{prop:LTS_l0}
If $\{\hat{\mathcal{V}},\hat{\mathbf{O}}\}$ minimizes
\eqref{eq:cost_l0unconstr} with $\lambda_0$ chosen such that
$\|\hat{\mathbf{O}}\|_0=N-\nu$, then
$\hat{\mathcal{V}}_{LTS}=\hat{\mathcal{V}}$.
\end{proposition}
\begin{IEEEproof}
Given $\lambda_0$ such that $\|\hat{\bbO}\|_0=N-\nu$, 
the goal is to characterize $\hat{\mathcal{V}}$ as well as 
the positions and values of the nonzero rows of $\hat{\bbO}$. Note 
that because $\|\hat{\bbO}\|_0=N-\nu$, the last term in the cost of
\eqref{eq:cost_l0unconstr} is constant, hence inconsequential to the 
minimization. Upon defining $\hat{\bbr}_n:=\bbx_n-\hat\bbm -\hat{\bbU}\hat{\bbs}_n$, it is not 
hard to see from the optimality conditions that the rows of $\hat{\bbO}$ satisfy
\begin{equation}\label{eq:min_oi}
\hat{\bbo}_n=\left\{\begin{array}{ccc}\mathbf{0}_p,&&\|\hat{\bbr}_{n}\|_2\leq \sqrt{\lambda_0}\\
\hat{\bbr}_{n},&&\|\hat{\bbr}_{n}\|_2>
\sqrt{\lambda_0}\end{array}\right.,\quad n=1,\ldots,N.
\end{equation}
This is intuitive, since for those nonzero $\hat{\bbo}_n$ the best
 thing to do in 
terms of minimizing the overall cost is to set $\hat{\bbo}_n=\hat{\bbr}_n$, and 
thus null the corresponding squared-residual terms in 
\eqref{eq:cost_l0unconstr}. In conclusion, for the chosen value of $\lambda_0$ 
it holds that $N-\nu$ squared residuals effectively do not contribute to the 
cost in \eqref{eq:cost_l0unconstr}. 

To determine $\hat{\mathcal{V}}$ and the row support of 
$\hat{\bbO}$, one alternative is to exhaustively test all 
${{N}\choose{N-\nu}}={{N}\choose{\nu}}$ admissible row-support combinations. For each one of these 
combinations (indexed by $j$), let 
$\mathcal{S}_j\subset\{1,\ldots,N\}$ be the index set describing the row support of
 $\hat{\bbO}^{(j)}$, i.e., $\hat{\bbo}_{n}^{(j)}\neq \mathbf{0}_p$ if and only if 
$n\in\mathcal{S}_j$; and $|\mathcal{S}_j|=N-\nu$. By virtue
 of \eqref{eq:min_oi}, the corresponding candidate $\hat{\mathcal{V}}^{(j)}$ 
solves $\min_{\mathcal{V}}\sum_{n\in\mathcal{S}_j}r_{n}^2(\mathcal{V})$
 subject to $\mathbf{U}^\prime\mathbf{U}=\mathbf{I}_q$,
while $\hat{\mathcal{V}}$ is the one among all $\{\hat{\mathcal{V}}^{(j)}\}$ that yields the least 
cost. Recognizing the aforementioned solution procedure as the one for LTS
PCA outlined  under Property \ref{prope:existence_LTS}, it follows that 
$\hat{\mathcal{V}}_{LTS}=\hat{\mathcal{V}}$.
\end{IEEEproof}
The importance of Proposition \ref{prop:LTS_l0} is threefold.
First, it formally justifies model \eqref{eq:model_outliers}
and its estimator \eqref{eq:cost_l0unconstr} for robust PCA, in
light of the well documented merits of LTS~\cite{RL87}. Second,
it further solidifies the connection between sparsity-aware
learning and robust estimation. Third, problem
\eqref{eq:cost_l0unconstr} lends itself naturally to efficient
(approximate) solvers based on convex relaxation, the subject
dealt with next.

% % % % % % % % % % % % % % % % % % % % % % % % % % % % % % % % % % % % % % % %
%                         Section III                                         %
% % % % % % % % % % % % % % % % % % % % % % % % % % % % % % % % % % % % % % % %

\vspace{-0.2cm}
\section{Sparsity-Controlling Outlier Rejection}\label{sec:spacor}

Recall that the row-wise $\ell_2$-norm sum
$\|\mathbf{B}\|_{2,r}:=\sum_{n=1}^N\|\mathbf{b}_n\|_2$ of
matrix
$\mathbf{B}:=[\mathbf{b}_1,\ldots,\mathbf{b}_N]^\prime\in\mathbb{R}^{N\times
p}$ is the closest convex approximation of $\|\mathbf{B}\|_0$.
This property motivates relaxing problem
\eqref{eq:cost_l0unconstr} to
\begin{equation}\label{eq:cost_l1unconstr}
\min_{\mathcal{V},\mathbf{O}}
\|\mathbf{X}-\mathbf{1}_N\mathbf{m}^\prime-
\mathbf{S}\mathbf{U}^\prime-\mathbf{O}\|_F^2
+\lambda_2\|\mathbf{O}\|_{2,r}
,\quad\textrm{s. to }\mathbf{U}^\prime\mathbf{U}=\mathbf{I}_q.
\end{equation}
The nondifferentiable $\ell_2$-norm regularization term
encourages row-wise (vector) sparsity on the estimator of
$\mathbf{O}$, a property that has been exploited in diverse
problems in engineering, statistics, and machine
learning~\cite{HTF09}. A noteworthy representative is the group
Lasso~\cite{yl06grouplasso}, a popular tool for joint
estimation and selection of grouped variables in linear
regression. 

It is pertinent to ponder on whether problem
\eqref{eq:cost_l1unconstr} still has the potential of providing
robust estimates $\hat{\mathcal{V}}$ in the presence of
outliers. The answer is positive, since it is shown in the Appendix
that 
%for a specific value of $\lambda_2$, 
\eqref{eq:cost_l1unconstr} is equivalent to an M-type estimator
\begin{equation}\label{eq:variational_w_rho}
\min_{\mathcal{V}}\sum_{n=1}^{N}\rho_v(\mathbf{x}_n-\mathbf{m}-\mathbf{U}
\mathbf{s}_n),\quad\textrm{s. to }\mathbf{U}^\prime\mathbf{U}=\mathbf{I}_q
\end{equation}
where $\rho_v:\mathbb{R}^p\to\mathbb{R}$ is a vector extension
to Huber's convex loss function~\cite{hr09}; see
also~\cite{kgg10}, and
\begin{equation}\label{eq:rho_def}
\rho_v(\mathbf{r}):=\left\{\begin{array}{ccc}\|\mathbf{r}\|_2^2,&&\|\mathbf{r}\|_2\leq\lambda_2/2\\
\lambda_2\|\mathbf{r}\|_2-\lambda_2^2/4,&&\|\mathbf{r}\|_2>\lambda_2/2\end{array}\right..
\end{equation}

M-type estimators (including Huber's) adopt a fortiori an 
$\epsilon$-contaminated probability distribution for the outliers, and rely
on minimizing the \textit{asymptotic} variance of the resultant estimator for 
the least favorable distribution
of the $\epsilon$-contaminated class (asymptotic min-max approach)~\cite{hr09}.
 The assumed degree of contamination
specifies the tuning parameter $\lambda_2$ in \eqref{eq:rho_def} (and thus the 
threshold for deciding
 the outliers in M-estimators). In contrast, the present approach is universal 
in the sense that it is not confined to any assumed class
of outlier distributions, and can afford a data-driven selection of the tuning 
parameter. In a nutshell,
M-estimators can be viewed as a special case of the present formulation only 
for a specific choice of $\lambda_2$, which is not obtained via a data-driven 
approach, but from distributional assumptions instead.

All in all, the sparsity-controlling role of the tuning parameter
$\lambda_2\geq 0$ in \eqref{eq:cost_l1unconstr} is central, since
model \eqref{eq:model_outliers} and the equivalence of \eqref{eq:cost_l1unconstr} with
\eqref{eq:variational_w_rho} suggest that $\lambda_2$ is a 
robustness-controlling 
constant. Data-driven approaches to
select $\lambda_2$ are described in detail under Section \ref{ssec:params}.
Before dwelling into algorithmic issues to solve \eqref{eq:cost_l1unconstr},
a couple of remarks are in order.

\begin{remark}[$\ell_1$-norm regularization for entry-wise outliers]\label{rem:l_1}
In computer vision applications where robust PCA schemes are
particularly attractive, one may not wish to discard the entire
(vectorized) images $\mathbf{x}_n$, but only specific pixels
deemed as outliers~\cite{dlt03}. This can be accomplished by
replacing $\|\mathbf{O}\|_{2,r}$ in \eqref{eq:cost_l1unconstr}
with $\|\mathbf{O}\|_{1}:=\sum_{n=1}^N\|\mathbf{o}_n\|_1$, a
Lasso-type regularization that encourages entry-wise sparsity
in $\hat{\mathbf{O}}$. 
\end{remark}

\begin{remark}[Outlier rejection]\label{rem:outlier_rejection}
From the equivalence
between problems \eqref{eq:cost_l1unconstr} and
\eqref{eq:variational_w_rho}, it follows that those data points
$\mathbf{x}_n$ deemed as containing outliers
$(\hat{\mathbf{o}}_n\neq\mathbf{0}_p)$ are not completely
discarded from the estimation process. Instead, their effect is
downweighted as per Huber's loss function [cf.
\eqref{eq:rho_def}]. Nevertheless, explicitly accounting for
the outliers in $\hat{\mathbf{O}}$ provides the means of
identifying and removing the contaminated data altogether, and
thus possibly re-running PCA on the outlier-free data.
\end{remark}

% % % % % % % % % % % % % % % % % % % % % % % % % % % % % % % % % % % % % % % %
%                         Subsection III-A                                    %
% % % % % % % % % % % % % % % % % % % % % % % % % % % % % % % % % % % % % % % %

\subsection{Solving the relaxed problem}\label{ssec:solve_l1unconstr}

To optimize \eqref{eq:cost_l1unconstr} iteratively for a given value of 
$\lambda_2$, an
alternating minimization (AM) algorithm is adopted which
cyclically updates
$\mathbf{m}(k)\to\mathbf{S}(k)\to\mathbf{U}(k)\to\mathbf{O}(k)$
per iteration $k=1,2,\ldots$. AM algorithms are also known as 
block-coordinate-descent 
methods in the optimization parlance; see e.g.,~\cite{Bertsekas_Book_Nonlinear,Tseng_cd_2001}. 
To update each of the variable
groups, \eqref{eq:cost_l1unconstr} is minimized while fixing
the rest of the variables to their most up-to-date values. While the overall
problem \eqref{eq:cost_l1unconstr} is not jointly convex with respect
to (w.r.t.) $\{\bbS,\bbU,\bbO,\bbm\}$, fixing all but one of the variable groups 
yields subproblems that are efficiently solved, 
and attain a unique solution. %per iteration.

Towards deriving the updates at iteration $k$ and arriving at the desired algorithm,
note first that the mean
update is $\mathbf{m}(k)=(\mathbf{X}-\mathbf{O}(k))^\prime
\mathbf{1}_N/N$. 
Next, form the centered and outlier-compensated data matrix
$\mathbf{X}_o(k):=\mathbf{X}-\mathbf{1}_N\mathbf{m}(k)^\prime-\mathbf{O}(k-1)$.
The principal components are readily given by
\begin{equation*}
\mathbf{S}(k)=\arg\min_{\bbS}\|\mathbf{X}_o(k)-\bbS\mathbf{U}(k-1)^{\prime}
\|_F^2=\mathbf{X}_o(k)\mathbf{U}(k-1).
\end{equation*}
Continuing the cycle, $\mathbf{U}(k)$ solves
\begin{equation*}
\min_{\mathbf{U}}
\|\mathbf{X}_o(k)-
\mathbf{S}(k)\mathbf{U}^\prime\|_F^2,\quad\textrm{s. to 
}\mathbf{U}^\prime\mathbf{U}=\mathbf{I}_q
\end{equation*}
a constrained LS problem also known as 
reduced-rank \textit{Procrustes rotation}~\cite{zouPCA}. The
minimizer is given in analytical form in terms of the left and right singular
vectors of $\mathbf{X}_o^\prime(k)\mathbf{S}(k)$~\cite[Thm.
4]{zouPCA}. In detail, one computes the SVD of
$\mathbf{X}_o^\prime(k)\mathbf{S}(k)=\mathbf{L}(k)\mathbf{D}(k)\mathbf{R}^\prime(k)$
and updates $\mathbf{U}(k)=\mathbf{L}(k)\mathbf{R}^\prime(k)$.
Next, the minimization of \eqref{eq:cost_l1unconstr} w.r.t. 
$\mathbf{O}$ is an orthonormal group Lasso problem.
As such, it decouples across rows $\mathbf{o}_n$
giving rise to $N$ $\ell_2$-norm regularized subproblems, namely
\begin{equation*}
\bbo_n(k)=\arg\min_{\bbo}
\|\mathbf{r}_{n}(k)-\bbo\|_2^2
+\lambda_2\|\bbo\|_{2},\:\quad n=1,\ldots,N
\end{equation*}
where $\mathbf{r}_{n}(k):=\mathbf{x}_n-\mathbf{m}(k)-\mathbf{U}(k)\mathbf{s}_n(k)$.
The
respective solutions are given by (see e.g.,~\cite{puig_hero})
\begin{equation}\label{eq:vector_soft_t}
\mathbf{o}_n(k)=\frac{\mathbf{r}_{n}(k)(\|\mathbf{r}_{n}(k)\|_2-\lambda_2/2)_+}{\|\mathbf{r}_{n}(k)\|_2},
\:\quad n=1,\ldots,N
\end{equation}
where $(\cdot)_+:=\max(\cdot,0)$. For notational convenience, these $N$ parallel vector
soft-thresholded updates are denoted as 
$\bbO(k)=\mathcal{S}\left[\mathbf{X}-\mathbf{1}_N\mathbf{m}^\prime(k-1)
-\mathbf{S}(k)\mathbf{U}^\prime(k),(\lambda_2/2)\mathbf{I}_N\right]$
under Algorithm \ref{table: AM_solver}, where the thresholding operator $\mathcal{S}$
sets the entire outlier vector $\bbo_n(k)$ to zero whenever $\|\bbr_n(k)\|_2$ does not exceed
$\lambda_2/2$, in par with the group sparsifying
property of group Lasso. Interestingly, this is the same rule
used to decide if datum $\bbx_n$ is deemed an outlier, 
in the equivalent formulation \eqref{eq:variational_w_rho} which involves Huber's loss function.
Whenever an
$\ell_1$-norm regularizer is adopted as discussed in Remark \ref{rem:l_1},
the only difference is that
updates \eqref{eq:vector_soft_t} boil down to soft-thresholding the scalar
entries of $\bbr_n(k)$.

The entire AM solver is tabulated under Algorithm \ref{table:
AM_solver}, indicating also the recommended initialization. Algorithm \ref{table:
AM_solver} is conceptually interesting, since it
explicitly reveals the intertwining between the outlier
identification process, and the PCA low-rank model fitting
based on the outlier compensated data $\mathbf{X}_o(k)$.

The AM solver is also computationally efficient. 
Computing the $N\times q$ matrix 
$\mathbf{S}(k)=\mathbf{X}_o(k)\mathbf{U}(k-1)$ requires $Npq$ operations
per iteration, and equally costly is 
to obtain $\mathbf{X}_o^\prime(k)\mathbf{S}(k)\in\mathbb{R}^{p\times q}$.
The cost of computing the SVD of $\mathbf{X}_o^\prime(k)\mathbf{S}(k)$ is of
order $\mathcal{O}(pq^2)$, while the rest of the operations including
the row-wise soft-thresholdings to yield $\bbO(k)$ 
are linear in both $N$ and $p$. In summary,
the total cost of Algorithm \ref{table: AM_solver} is roughly 
$k_{\max}\mathcal{O}(Np+pq^2)$, where $k_{\max}$ is the number of iterations
required for convergence (typically $k_{\max}=5$ to $10$ iterations suffice).
Because $q\leq p$ is typically small, Algorithm \ref{table: AM_solver} is
attractive computationally both under the classic setting where $N>p$, and $p$
is not large; 
as well as in high-dimensional data settings where $p\gg N$, a situation 
typically arising e.g., in microarray data analysis.

Because each of the optimization problems in the per-iteration cycles has a 
unique minimizer, and the
nondifferentiable regularization only affects one of the
variable groups $(\mathbf{O})$, the general  results of~\cite{Tseng_cd_2001} 
apply to establish convergence of Algorithm \ref{table: AM_solver} as follows.

\begin{proposition}\label{prop:conv_alg_1}
As $k\to\infty$, the iterates generated by Algorithm
\ref{table: AM_solver} converge to a stationary point of
\eqref{eq:cost_l1unconstr}.
\end{proposition}

\begin{algorithm}[t]
\caption{: Batch robust PCA solver}
\small{
\begin{algorithmic}
    \STATE Set $\mathbf{U}(0)=\mathbf{I}_p(:,1:q)$ and $\mathbf{O}(0)=\mathbf{0}_{N\times p}$.
    \FOR {$k=1,2,\ldots$}
    	\STATE Update $\mathbf{m}(k)=(\mathbf{X}-\mathbf{O}(k-1))^\prime \mathbf{1}_N/N$.
        \STATE Form $\mathbf{X}_o(k)=\mathbf{X}-\mathbf{1}_N\mathbf{m}^\prime(k)-\mathbf{O}(k-1)$.
        \STATE Update $\mathbf{S}(k)=\mathbf{X}_o(k)\mathbf{U}(k-1)$.	
	    \STATE Obtain
$\mathbf{L}(k)\mathbf{D}(k)\mathbf{R}(k)^\prime=
\textrm{svd}[\mathbf{X}_o^\prime(k)\mathbf{S}(k)]$ and update 
$\mathbf{U}(k)=\mathbf{L}(k)\mathbf{R}^\prime(k)$.   	
        \STATE Update $\mathbf{O}(k)=\mathcal{S}\left[\mathbf{X}-\mathbf{1}_N\mathbf{m}^\prime(k)
        -\mathbf{S}(k)\mathbf{U}^\prime(k),(\lambda_2/2)\mathbf{I}_N\right].$
    \ENDFOR
	\end{algorithmic}} \label{table: AM_solver}
\end{algorithm}

% % % % % % % % % % % % % % % % % % % % % % % % % % % % % % % % % % % % % % % %
%                         Subsection III-B                                    %
% % % % % % % % % % % % % % % % % % % % % % % % % % % % % % % % % % % % % % % %

\subsection{Selection of $\lambda_2$: robustification paths}\label{ssec:params}

Selecting $\lambda_2$ controls the number of outliers rejected.
But this choice is challenging because existing techniques such
as cross-validation are not effective when outliers are
present~\cite{RL87}. To this end, systematic data-driven approaches were 
devised in~\cite{uspacor}, which e.g., require a rough estimate of the 
percentage
of outliers, or, robust estimates $\hat{\sigma}_{e}^2$ of the
nominal noise variance that can be obtained using median
absolute deviation (MAD) schemes~\cite{hr09}. These approaches can be adapted
to the robust PCA setting considered here, and
leverage the \textit{robustification paths} of
(group-)Lasso solutions [cf. \eqref{eq:cost_l1unconstr}], which are defined as 
the solution paths corresponding
to $\|\hat{\bbo}_n\|_2,$ $n=1,\ldots,N$, for all values of
$\lambda_2$.  As $\lambda_2$
decreases, more vectors $\hat{\bbo}_n$ enter
the model signifying that more of the training data are deemed to contain 
outliers.

Consider then a grid of $G_\lambda$ values of $\lambda_2$ in the interval
$[\lambda_{\min},\lambda_{\max}]$, evenly spaced on a logarithmic
scale. Typically,
$\lambda_{\max}$ is chosen as the minimum $\lambda_2$ value such that
$\hat{\mathbf{O}}\neq\mathbf{0}_{N\times p}$,
while 
$\lambda_{\min}=\epsilon\lambda_{\max}$ with
$\epsilon=10^{-4}$, say. 
Because Algorithm \ref{table: AM_solver} converges quite fast, 
\eqref{eq:cost_l1unconstr}
can be efficiently solved over the grid of $G_\lambda$ values for
$\lambda_2$. In the order of hundreds of grid points can be easily handled
by initializing each instance of Algorithm 1 (per value of $\lambda_2$) 
using \textit{warm starts}~\cite{HTF09}. This means
that multiple instances of \eqref{eq:cost_l1unconstr} are solved for a sequence of 
decreasing $\lambda_2$ values, and the initialization of Algorithm \ref{table: AM_solver} per grid point
corresponds to the solution obtained for the immediately preceding value
of $\lambda_2$ in the grid. For sufficiently close values of $\lambda_2$,
one expects that the respective solutions will also be close (the row
support of $\hat{\mathbf{O}}$ will most likely not change), and hence Algorithm
1 will converge after few iterations.  

Based on the $G_\lambda$ samples of the robustification paths and the prior knowledge
available on the outlier model \eqref{eq:model_outliers}, 
a couple of alternatives are also possible for selecting the `best'
value of $\lambda_2$ in the grid. A comprehensive survey of options
can be found in~\cite{uspacor}.

\noindent\textit{Number of outliers is known:} By direct
inspection of the robustification paths one can determine the range
of values for $\lambda_2$, such that the number of nonzero rows in
$\hat{\mathbf{O}}$ equals the known number of outliers sought. Zooming-in
to the interval of interest, and after discarding the identified
outliers, $K$-fold cross-validation methods can be applied to determine the
`best' $\lambda_2^{\ast}$.

\noindent\textit{Nominal noise covariance matrix is known:} Given $\bm\Sigma_e:=
E[\bbe_n\bbe^{\prime}_n]$, one can
proceed as follows. Consider the estimates $\hat{\mathcal{V}}_g$
obtained using \eqref{eq:cost_l1unconstr} after
sampling the robustification path for each point
$\{\lambda_{2,g}\}_{g=1}^{G}$. Next, pre-whiten those residuals
corresponding to training data not deemed as containing outliers; i.e., form
$\hat{\mathcal{R}}_g:=\{\bar{\bbr}_{n,g}=\bm\Sigma_e^{-1/2}(\mathbf{x}_n-\hat{\bbm}_g-\hat{\bbU}_g\hat{\bbs}_{n,g}):
n\textrm{ s. to }\hat{\bbo}_n= \mathbf{0}\}$, and find
the sample covariance matrices $\{\hat{\bm\Sigma}_{\bar{r},g}\}_{g=1}^G$.
The winner $\lambda_2^{\ast}:=\lambda_{2,g^{\ast}}$ corresponds to the
grid point minimizing an absolute variance deviation criterion, namely
$g^{\ast}:=\arg\min_{g}|\textrm{tr}[\hat{\bm\Sigma}_{\bar{r},g}]-p|$.

% % % % % % % % % % % % % % % % % % % % % % % % % % % % % % % % % % % % % % % %
%                         Subsection III-C                                    %
% % % % % % % % % % % % % % % % % % % % % % % % % % % % % % % % % % % % % % % %

\subsection{Connections with robust linear regression, dictionary learning, and
clustering}\label{ssec:connections}

Previous efforts towards robustifying linear regression have
pointed out the equivalence between M-type estimators and
$\ell_1$-norm regularized regression~\cite{fuchs99}, and capitalized
on this neat connection under a Bayesian framework~\cite{rao_robust}. However,
they have not recognized the link to LTS via convex
relaxation of the $\ell_0$-norm in \eqref{eq:cost_l0unconstr}. The 
treatment here
goes beyond linear regression by considering the PCA framework, which
entails a more challenging bilinear factor analysis model.
Linear regression is subsumed as a special case, when matrix
$\mathbf{U}$ is not necessarily tall but \textit{assumed
known}, while $\mathbf{s}_n=\mathbf{s},$ $\forall\:n=1,\ldots,N$.

As an alternative to PCA, it is possible to device
dimensionality reduction schemes when the data
admit a sparse representation over a perhaps \textit{unknown} basis.
Such sparse representations
comprise only a few elements (atoms) of the overcomplete basis (a.k.a.
dictionary) to reconstruct the original data record. Thus, each
datum is represented by a coefficient vector whose effective
dimensionality (number of nonzero coefficients) is smaller than that
of the original data vector. Recently, the \textit{dictionary learning}
paradigm offers techniques to design a dictionary over which the data
assume a sparse representation; see e.g.,~\cite{Dictionary_learning_SP_mag_10}
for a tutorial treatment. Dictionary learning schemes are
flexible, in the sense that they utilize training data to learn an
appropriate overcomplete basis customized for the data at
hand~\cite{mairal10,Dictionary_learning_SP_mag_10}. 

However, as in PCA the criteria adopted typically rely on a squared-error loss 
function as
a measure of fit, which is known to be very sensitive to 
outliers~\cite{RL87,hr09}. Interestingly, one can conceivably think of 
robustifying dictionary learning via minor modifications to the framework 
described so far. For instance, with the same matrix notation used in e.g., 
\eqref{eq:cost_l0unconstr}, one seeks to minimize
\begin{equation}\label{eq:cost_dictionary_learning}
\min_{\mathcal{V},\mathbf{O}}
\|\mathbf{X}-\mathbf{S}\mathbf{U}^\prime-\mathbf{O}\|_F^2
+\lambda_1\|\mathbf{S}\|_1+\lambda_2\|\mathbf{O}\|_{2,r}.
\end{equation}
Different from the low-rank outlier-aware model adopted for PCA [cf.
\eqref{eq:model_outliers}], here the dictionary
$\mathbf{U}\in\mathbb{R}^{p\times q}$ is fat $(q\gg p)$, with column vectors
that are no longer orthogonal but still constrained to have unit 
$\ell_2$-norm. (This constraint is left implicit in 
\eqref{eq:cost_dictionary_learning} for simplicity.)
Moreover, one seeks a sparse vector $\mathbf{s}_n$ to represent 
each datum $\mathbf{x}_n$, in terms of a few atoms of the 
learnt dictionary $\hat{\mathbf{U}}$. This is why \eqref{eq:cost_dictionary_learning}
includes an additional sparsity-promoting $\ell_1$-norm regularization on 
$\mathbf{S}$, 
that is not present in \eqref{eq:cost_l1unconstr}.
Sparsity is thus present both in the representation coefficients $\mathbf{S}$, 
as well as in the outliers $\mathbf{O}$. 

Finally, it is shown here that a generative data model for K-means 
clustering~\cite{HTF09} 
can share striking similarities with the bilinear model \eqref{eq:factor_model}. 
Consequently, the sparsity-controlling estimator \eqref{eq:cost_l1unconstr} can be adapted 
to robustify the K-means clustering task too~\cite{pf_vk_gg_clustering}. 
Consider for instance that the data in $\mathcal{T}_x$ come from $q$ clusters, each
of which is represented by a centroid $\mathbf{u}_i\in\mathbb{R}^p$, $i=1,\ldots,q$. 
Moreover, for each input vector $\mathbf{x}_n$, 
K-means introduces the unknown membership variables $s_{ni}\in\{0,1\}$, 
$i=1,\ldots,q$, where $s_{ni}=1$
whenever $\bbx_n$ comes from cluster $i$, and $s_{ni}=0$ otherwise. Typically, the 
membership variables are also constrained to satisfy $\sum_{n=1}^Ns_{ni}>0$ 
$\forall\:i$
(no empty clusters), and $\sum_{i=1}^qs_{ni}=1$ $\forall\:n$ (single cluster 
membership). Upon
defining $\mathbf{U}:=[\mathbf{u}_1,\ldots,\mathbf{u}_q]\in\mathbb{R}^{p\times q}$
and the membership vectors $\mathbf{s}_n:=[s_{n1},\ldots,s_{nq}]^\prime\in\mathbb{R}^q$, 
a pertinent model for hard K-means clustering assumes that input vectors can 
be expressed
as $\mathbf{x}_n=\mathbf{U}\mathbf{s}_n+\mathbf{e}_n+\mathbf{o}_n$, where $\mathbf{e}_n$
and $\mathbf{o}_n$ are as in \eqref{eq:model_outliers}. 
%Without considering 
%outliers, a similar bilinear model was advocated 
%by~\cite{nikos_biclustering}, 
%as the conceptual building block for more general multilinear decompositions 
%suitable for bi-clustering tasks. 
Because the 
aforementioned constraints imply $\|\mathbf{s}_n\|_0=
\|\mathbf{s}_n\|_1=1$ $\forall\:n$, 
if $\mathbf{x}_n$ belongs to cluster $i$, then $s_{ni}=1$ and in the absence
of outliers one effectively has $\mathbf{x}_n=\mathbf{u}_i+\mathbf{e}_n$. 
Based on this data model, a natural approach towards robustifying K-means 
clustering solves~\cite{pf_vk_gg_clustering}
\begin{equation}\label{eq:cost_clustering}
\min_{\mathcal{V},\mathbf{O}}
\|\mathbf{X}-
\mathbf{S}\mathbf{U}^\prime-\mathbf{O}\|_F^2
+\lambda_2\|\mathbf{O}\|_{2,r},\quad\:
\textrm{s. to }s_{ni}\in\{0,1\},\:\sum_{n=1}^Ns_{ni}>0,\:\sum_{i=1}^q 
s_{ni}=1.
\end{equation}
Recall that in the robust PCA estimator \eqref{eq:cost_l1unconstr}, the 
subspace matrix is required to be orthonormal and the principal components are 
unrestrained. In the clustering context however, the centroid columns of 
$\mathbf{U}$ are free optimization variables, whereas the cluster membership 
variables adhere to the constraints in \eqref{eq:cost_clustering}. Suitable 
relaxations to tackle the NP-hard problem \eqref{eq:cost_clustering} have been 
investigated in~\cite{pf_vk_gg_clustering}.

% % % % % % % % % % % % % % % % % % % % % % % % % % % % % % % % % % % % % % % %
%                         Section IV                                          %
% % % % % % % % % % % % % % % % % % % % % % % % % % % % % % % % % % % % % % % %

\section{Further Algorithmic Issues}\label{sec:algo_issues}

% % % % % % % % % % % % % % % % % % % % % % % % % % % % % % % % % % % % % % % %
%                         Subsection IV-A                                     %
% % % % % % % % % % % % % % % % % % % % % % % % % % % % % % % % % % % % % % % %

\subsection{Bias reduction through nonconvex regularization}\label{ssec:nonconvex}

Instead of substituting $\|\mathbf{O}\|_0$ in \eqref{eq:cost_l0unconstr}
by its closest convex approximation, namely
$\|\mathbf{O}\|_{2,r}$, letting the surrogate function to be
nonconvex can yield tighter approximations, and improve the statistical 
properties of the estimator. In
rank minimization problems for instance, the logarithm of the
determinant of the unknown matrix has been proposed as 
a smooth surrogate to the rank~\cite{fazel_log_det}; an alternative to the 
 convex nuclear norm in e.g.,~\cite{Recht_SIAM_2010}. 
Nonconvex penalties such as the
smoothly clipped absolute deviation (SCAD)
have been also adopted to reduce bias~\cite{scad}, present in uniformly weighted
$\ell_1$-norm regularized estimators such as
\eqref{eq:cost_l1unconstr}~\cite[p. 92]{HTF09}. 
In the context of sparse signal reconstruction, the
$\ell_0$-norm of a vector 
was surrogated in~\cite{candes_l0_surrogate} by the logarithm
of the geometric mean of its elements; see also~\cite{nacho_universal}.

Building on this last idea, 
consider approximating \eqref{eq:cost_l0unconstr} by the
\textit{nonconvex} formulation
\begin{equation}\label{eq:cost_nonconvex}
\min_{\mathcal{V},\mathbf{O}}
\|\mathbf{X}-\mathbf{1}_N\mathbf{m}^\prime-
\mathbf{S}\mathbf{U}^\prime-\mathbf{O}\|_F^2
+\lambda_0\sum_{n=1}^N\log(\|\mathbf{o}_n\|_2+\delta),
\quad\textrm{s. to }\mathbf{U}^\prime\mathbf{U}=\mathbf{I}_q
\end{equation}
where the small positive constant $\delta$ is introduced to avoid numerical
instability. Since the surrogate term in \eqref{eq:cost_nonconvex} is concave, the overall 
minimization problem is nonconvex and admittedly more complex to solve
than \eqref{eq:cost_l1unconstr}. Local methods based on iterative 
linearization of
$\log(\|\mathbf{o}_n\|_2+\delta)$ around the current iterate
$\mathbf{o}_n(k)$, are adopted to minimize
\eqref{eq:cost_nonconvex}. Skipping details that
can be found in~\cite{kgg10}, application of the majorization-minimization 
technique 
to \eqref{eq:cost_nonconvex} leads to an 
iteratively-reweighted version of \eqref{eq:cost_l1unconstr},
whereby $\lambda_2\leftarrow\lambda_0 w_n(k)$ is used for updating
$\mathbf{o}_n(k)$ in Algorithm \ref{table: AM_solver}. Specifically, per 
$k=1,2,\ldots$ one updates
\begin{equation*}
\mathbf{O}(k)=\mathcal{S}\left[\mathbf{X}-\mathbf{1}_N\mathbf{m}^\prime(k-1)
        -\mathbf{S}(k)\mathbf{U}^\prime(k),(\lambda_0/2)\textrm{diag}(w_1(k),\ldots,w_N(k))\right]
\end{equation*}
where the weights are given by
$w_n(k)=\left(\|\mathbf{o}_n(k-1)\|_2+\delta\right)^{-1},$ $n=1,\ldots,N.$
Note that the thresholds vary both across rows (indexed by $n$), and across 
iterations.
If the value of $\|\mathbf{o}_n(k-1)\|_2$ is small,
then in the next iteration the regularization
term $\lambda_0 w_n(k)\|\mathbf{o}_n\|_2$ has a large weight, thus
promoting shrinkage of that entire row vector to zero. If 
$\|\mathbf{o}_n(k-1)\|_2$ is large, the cost in the next
iteration downweighs the regularization, and places more
importance to the LS component of the fit.

All in all, the idea is to start from the solution
of \eqref{eq:cost_l1unconstr} for the `best' $\lambda_2$, which is obtained 
using Algorithm \ref{table: AM_solver}. This initial estimate is
refined after runnning a few 
iterations of the iteratively-reweighted counterpart to Algorithm \ref{table: AM_solver}.
Extensive numerical
tests suggest that even a couple iterations of this second
stage refinement suffices to yield improved estimates
$\hat{\mathcal{V}}$, in comparison to those obtained from
\eqref{eq:cost_l1unconstr}. The improvements can be leveraged
to bias reduction -- and its positive effect with regards to outlier support 
estimation -- also achieved by similar \textit{weighted}
norm regularizers proposed for linear
regression~\cite[p. 92]{HTF09}.

% % % % % % % % % % % % % % % % % % % % % % % % % % % % % % % % % % % % % % % %
%                         Subsection IV-B                                     %
% % % % % % % % % % % % % % % % % % % % % % % % % % % % % % % % % % % % % % % %

\subsection{Automatic rank determination: from nuclear- to Frobenius-norm 
regularization}\label{ssec:Frob}

Recall that $q\leq p$ is the 
dimensionality of the subspace where the outlier-free data 
\eqref{eq:factor_model} are assumed to live 
in, or equivalently, $q=\textrm{rank}[\mathbf{Y}]$ in the absence of noise. So 
far, $q$  
was assumed known and fixed. This is reasonable in e.g., 
compression/quantization, where a target distortion-rate tradeoff 
dictates the maximum $q$. In other cases, the physics of the problem may 
render $q$ 
known. This is indeed the case in array processing for direction-of-arrival 
estimation, where $q$ is the dimensionality of the so-termed \textit{signal 
subspace}, and is given by the number of plane waves impinging on a uniform linear 
array; see e.g.,~\cite{yang95}. 

Other applications however, call for signal processing tools that can determine
 the `best' $q$, as well as robustly estimate the underlying low-dimensional 
subspace $\mathbf{U}$ from data $\mathbf{X}$. Noteworthy representatives for
this last 
kind of problems include unveiling traffic volume anomalies in large-scale 
networks~\cite{morteza_gonzalo_gg_asilomar11}, and automatic intrusion 
detection from video surveillance frames~\cite{dlt03,clmw09}, 
just to name a few. A related approach in this context is 
(stable) principal components pursuit (PCP)~\cite{zlwcm10,Outlier_pursuit}, 
which solves
\begin{equation}\label{eq:spcp}
\min_{\mathbf{L},\mathbf{O}}
\|\mathbf{X}-\mathbf{L}-\mathbf{O}\|_F^2
+\lambda_{\ast}\|\mathbf{L}\|_{\ast}+\lambda_2\|\mathbf{O}\|_{2,r}
\end{equation}
with the objective of reconstructing the low-rank matrix 
$\mathbf{L}\in\mathbb{R}^{N\times p}$, as well as the sparse matrix of 
outliers $\mathbf{O}$
 in the presence of dense
 noise with known variance.\footnote{Actually,~\cite{zlwcm10} considers
 entrywise outliers and adopts an $\ell_1$-norm regularization on
 $\mathbf{O}$.} Note 
that $\|\mathbf{L}\|_{\ast}$  denotes the 
matrix nuclear norm, defined as the sum of the singular values of 
$\mathbf{L}$. The same way that the $\ell_2$-norm regularization promotes 
sparsity in the rows of $\hat{\mathbf{O}}$, the nuclear norm encourages a
low-rank $\hat{\mathbf{L}}$ since it effects
sparsity in the vector of singular values of $\mathbf{L}$. Upon solving the 
convex optimization problem \eqref{eq:spcp}, it is possible to obtain 
$\hat{\mathbf{L}}=\hat{\mathbf{S}}\hat{\mathbf{U}}^\prime$ using the SVD. 
Interestingly, \eqref{eq:spcp} does not fix (or require the knowledge of) 
$\textrm{rank}[\mathbf{L}]$ a fortiori, but controls it through the tuning 
parameter $\lambda_{\ast}$. Adopting a Bayesian framework, a similar problem
was considered in~\cite{bayes_rpca}.

Instead of assuming that $q$ is known, suppose that only an upper
 bound $\bar{q}$ is given. Then, the class of feasible noise-free 
low-rank matrix 
components of $\mathbf{Y}$ in \eqref{eq:factor_model} admit a factorization  
$\mathbf{L}=\mathbf{S}\mathbf{U}^\prime$, where $\mathbf{S}$ and $\mathbf{U}$ 
are $N\times\bar{q}$ and $p\times\bar{q}$ matrices, respectively. Building on 
the ideas used in the context of finding minimum rank solutions of linear 
matrix equations~\cite{Recht_SIAM_2010}, a novel alternative approach to robustifying 
PCA is to solve 
\begin{equation}\label{eq:frob_reg}
\min_{\mathbf{U},\mathbf{S},\mathbf{O}}
\|\mathbf{X}-\mathbf{S}\mathbf{U}^\prime-\mathbf{O}\|_F^2
+\frac{\lambda_{\ast}}{2}(\|\mathbf{U}\|_{F}^2+\|\mathbf{S}\|_{F}^2)+\lambda_2
\|\mathbf{O}\|_{2,r}.
\end{equation}
Different from \eqref{eq:spcp} and \eqref{eq:cost_l1unconstr}, a 
Frobenius-norm regularization on both 
$\mathbf{U}$ and $\mathbf{S}$ is adopted to control the dimensionality of the 
estimated subspace $\hat{\mathbf{U}}$. Relative to \eqref{eq:cost_l1unconstr},
$\mathbf{U}$ in \eqref{eq:frob_reg} is not constrained to be orthonormal.
It is certainly possible to include the 
mean vector $\mathbf{m}$ in the cost of \eqref{eq:frob_reg}, as well as an $\ell_1$-norm 
regularization for entrywise outliers. The main motivation behind choosing the 
Frobenius-norm regularization comes from the equivalence 
of \eqref{eq:spcp} with \eqref{eq:frob_reg},
 as asserted in the ensuing result which 
 adapts~\cite[Lemma 5.1]{Recht_SIAM_2010} to the problem formulation 
considered here. 

\begin{lemma}\label{lemma:equivalence}
If $\{\hat{\mathbf{L}},\hat{\mathbf{O}}\}$ minimizes 
\eqref{eq:spcp} and $\textrm{rank}[\hat{\mathbf{L}}]\leq \bar{q}$, then
\eqref{eq:spcp} and \eqref{eq:frob_reg} are equivalent.
\end{lemma}
\begin{IEEEproof}
Because $\textrm{rank}[\hat{\mathbf{L}}]\leq \bar{q}$, the relevant feasible
 subset of 
\eqref{eq:spcp} can be re-parametrized as 
$\{\mathbf{S}\mathbf{U}^\prime,\mathbf{O}\}$, where $\mathbf{S}$ and 
$\mathbf{U}$ are $N\times\bar{q}$ and $p\times\bar{q}$ matrices, respectively. 
For every triplet $\{\mathbf{U},\mathbf{S},\mathbf{O}\}$ the objective of 
\eqref{eq:frob_reg} is no smaller than the one of \eqref{eq:spcp}, since
it holds that~\cite{Recht_SIAM_2010}
\begin{equation}\label{eq:nuclear_ min}
\|\mathbf{L}\|_{\ast}=\min_{\mathbf{U},\mathbf{S}}\frac{1}{2}(\|\mathbf{U}\|_{
F}^2+\|\mathbf{S}\|_{F}^2),\quad\textrm{s. to }\mathbf{L}=\mathbf{S}\mathbf{U}^\prime.
\end{equation}
One can  show that the gap between the 
objectives 
of \eqref{eq:spcp} and \eqref{eq:frob_reg} 
vanishes at 
$\mathbf{O}^{\ast}:=\hat{\mathbf{O}}$, 
$\mathbf{S}^{\ast}:=\mathbf{U}_L\bm\Sigma^{1/2}$, and 
$\mathbf{U}^{\ast}:=\mathbf{V}_L\bm\Sigma^{1/2}$; where 
$\hat{\mathbf{L}}=\mathbf{U}_L\bm\Sigma\mathbf{V}_L^\prime$ is the SVD of 
$\hat{\mathbf{L}}$. Therefore, from the previous arguments it follows that 
\eqref{eq:spcp} and \eqref{eq:frob_reg} attain the same global minimum 
objective, which completes the proof.
\end{IEEEproof}

Even though problem \eqref{eq:frob_reg} is nonconvex, the number of 
optimization variables is reduced from $2Np$ to $Np+(N+p)\bar{q}$, which 
becomes significant when $\bar{q}$ is in the order of a few dozens and both $N$
 and $p$ are large. Also note that the
 dominant $Np$-term in the variable count of \eqref{eq:frob_reg} is due to 
$\mathbf{O}$, which is sparse and can be efficiently handled. While the 
factorization $\mathbf{L}=\mathbf{S}\mathbf{U}^\prime$ could 
have also been introduced in \eqref{eq:spcp} to reduce the number of unknowns, 
the cost in \eqref{eq:frob_reg} is separable and much simpler to optimize using
 e.g., an AM solver comprising the iterations tabulated as Algorithm 
\ref{table: AM_solver_rank_control}. The 
decomposability of the Frobenius-norm regularizer has been recently exploited 
for parallel processing across multiple processors when solving large-scale 
matrix completion problems~\cite{Recht_Parallel_2011}, or to unveil 
network
anomalies~\cite{morteza_gonzalo_gg_asilomar11}.

\begin{algorithm}[t]
\caption{: Batch robust PCA solver with controllable rank} \small{
\begin{algorithmic}
    \STATE Set $\mathbf{O}(0)=\mathbf{0}_{N\times p}$, and randomly initialize $\mathbf{S}(0)$.
    \FOR {$k=1,2,\ldots$}
        \STATE Update $\mathbf{m}(k)=[\mathbf{X}-\mathbf{O}(k-1)]^\prime \mathbf{1}_N/N.$
        \STATE Form $\mathbf{X}_o(k)=\mathbf{X}-\mathbf{1}_N\mathbf{m}^\prime(k)-\mathbf{O}(k-1)$.
        \STATE Update $\mathbf{U}(k)=\mathbf{X}_o(k)^\prime\mathbf{S}(k-1)
                [\mathbf{S}^\prime(k-1)\mathbf{S}(k-1)+
                (\lambda_\ast/2)\mathbf{I}_{\bar{q}}]^{-1}$.
        \STATE Update 
                $\mathbf{S}(k)=\mathbf{X}_o(k)\mathbf{U}(k)
                        [\mathbf{U}^\prime(k)\mathbf{U}(k)+
                        (\lambda_\ast/2)\mathbf{I}_{\bar{q}}]^{-1}$.
        \STATE Update $\mathbf{O}(k)=\mathcal{S}\left[\mathbf{X}
        -\mathbf{S}(k)\mathbf{U}^\prime(k),\lambda_2/2\right].$
    \ENDFOR
	\end{algorithmic}} \label{table: AM_solver_rank_control}
\end{algorithm}

Because \eqref{eq:frob_reg} is a nonconvex optimization problem, most 
solvers one can think of will at most provide convergence guarantees to a 
stationary point that may not be globally optimum. Nevertheless, simulation 
results in Section \ref{sec:sims} demonstrate that Algorithm \ref{table: 
AM_solver_rank_control} is effective in providing good solutions most of the 
time, which is somehow expected since there is quite a bit of structure in 
\eqref{eq:frob_reg}. Formally, the next proposition adapted 
from~\cite[Prop. 1]{morteza_gonzalo_gg_asilomar11} provides a sufficient 
condition under which Algorithm \ref{table: AM_solver_rank_control} yields an 
optimal solution of \eqref{eq:spcp}. For a proof of a slightly more general 
result, see~\cite{morteza_gonzalo_gg_asilomar11}.

\begin{proposition}\label{prop:prop_equiv_frob}
If $\{\bar{\mathbf{U}},\bar{\mathbf{S}},\bar{\mathbf{O}}\}$ is a stationary 
point of \eqref{eq:frob_reg} and $\|\mathbf{X}-
\bar{\mathbf{S}}\bar{\mathbf{U}}'-\bar{\mathbf{O}}\|_2\leq \lambda_*/2$, 
then 
$\{\hat{\mathbf{L}}:=\bar{\mathbf{S}}\bar{\mathbf{U}}', 
\hat{\mathbf{O}}:=\bar{\mathbf{O}}\}$ is the optimal solution of 
\eqref{eq:spcp}.
\end{proposition}

% % % % % % % % % % % % % % % % % % % % % % % % % % % % % % % % % % % % % % % %
%                         Section V                                           %
% % % % % % % % % % % % % % % % % % % % % % % % % % % % % % % % % % % % % % % %

\section{Robust Subspace Tracking}\label{sec:online}

E-commerce and Internet-based retailing sites, the World Wide Web, and video
surveillance systems generate huge volumes of data, which far
outweigh the ability of modern computers to analyze them in
real time. Furthermore, data are generated sequentially in
time, which motivates updating previously obtained learning
results rather than re-computing new ones from scratch each
time a new datum becomes available. This calls for
low-complexity real-time (adaptive) algorithms for robust
subspace tracking. 

One possible adaptive counterpart to
\eqref{eq:cost_l1unconstr} is the exponentially-weighted LS
(EWLS) estimator found by
\begin{equation}\label{eq:EWLS}
\hspace{-0.10cm}\min_{\{\mathcal{V},\mathbf{O}\}}
\sum_{n=1}^{N}\beta^{N-n}\left[\|\mathbf{x}_{n}-\mathbf{m}-
\mathbf{U}\mathbf{s}_n-\mathbf{o}_n\|_2^2
+\lambda_2\|\mathbf{o}_n\|_2\right]
\end{equation}
where $\beta\in(0,1]$ is a forgetting factor. In this context, $n$ should be 
understood as a temporal variable, indexing the 
instants of data acquisition.  Note that in
forming the EWLS estimator \eqref{eq:EWLS} at time $N$, the
entire history of data $\{\mathbf{x}_n\}_{n=1}^N$ is
incorporated in the real-time estimation process. Whenever
$\beta<1$, past data are exponentially discarded thus enabling
operation in nonstationary environments. Adaptive estimation
of sparse signals has been considered in e.g.,~\cite{da_jn_gg_2010} 
and~\cite{mairal10}.

Towards deriving a real-time, computationally efficient, and
recursive (approximate) solver of \eqref{eq:EWLS}, an AM scheme
will be adopted in which iterations $k$ coincide with the time
scale $n=1,2,\ldots$ of data acquisition. Per time instant $n$,
a new datum $\mathbf{x}_n$ is drawn and the corresponding
pair of decision variables $\{\mathbf{s}(n),\mathbf{o}(n)\}$ are updated via
\begin{equation}\label{eq:sparse_coding}
\{\mathbf{s}(n),\mathbf{o}(n)\}:=\arg\min_{\{\mathbf{s},\mathbf{o}\}}
\|\mathbf{x}_{n}-\mathbf{m}(n-1)-
\mathbf{U}(n-1)\mathbf{s}-\mathbf{o}\|_2^2
+\lambda_2\|\mathbf{o}\|_2.
\end{equation}
As per \eqref{eq:sparse_coding}, only $\mathbf{o}(n)$ is
updated at time $n$, rather than the whole (growing with time)
matrix $\mathbf{O}$ that minimization of \eqref{eq:EWLS} would
dictate; see also~\cite{mairal10} for a similar approximation.  

Because \eqref{eq:sparse_coding} is a smooth optimization problem w.r.t. 
$\mathbf{s}$, from the first-order optimality condition the
principal component update is
$\mathbf{s}(n)=\mathbf{U}^\prime(n-1)[\mathbf{x}_n-\mathbf{m}(n-1)-\mathbf{o}(n
)]$. Interestingly, this resembles the projection approximation adopted
in~\cite{yang95}, and can only be evaluated after $\mathbf{o}(n)$ is obtained. 
To 
this end, plug $\mathbf{s}(n)$ in \eqref{eq:sparse_coding} to 
obtain $\mathbf{o}(n)$ via a particular instance of the group Lasso estimator
\begin{equation}\label{eq:sparse_coding_o}
\mathbf{o}(n)=\arg\min_{\mathbf{o}}
\|[\mathbf{I}_p-\mathbf{U}(n-1)\mathbf{U}^\prime(n-1)](\mathbf{x}_{n}-\mathbf{m
}(n-1)-\mathbf{o})\|_2^2
+\lambda_2\|\mathbf{o}\|_2
\end{equation}
with a single group of size equal to $p$. The cost in \eqref{eq:sparse_coding_o} 
is non-differentiable at the origin, and different from e.g., ridge 
regression, it does not admit a closed-form solution. Upon defining
\begin{align} 
\label{eq:H}\mathbf{H}(n):={}&2[\mathbf{I}_p-\mathbf{U}(n-1)\mathbf{U}^\prime(n-1)]^\prime
[\mathbf{I}_p-\mathbf{U}(n-1)\mathbf{U}^\prime(n-1)]\in\mathbb{R}^{p\times p}\\
\label{eq:g}\mathbf{g}(n):={}&-\mathbf{H}(n)[\mathbf{x}_{n}-\mathbf{m}(n-1)]\in\mathbb{R}^p
\end{align}
one can recognize 
$\eqref{eq:sparse_coding_o}$ as the multidimensional shrinkage-thresholding 
operator $\mathcal{T}_{\mathbf{H}(n),\lambda_2}(\mathbf{g}(n))$ introduced 
in~\cite{puig_hero}. In particular, as per~\cite[Corollary 2]{puig_hero} it 
follows that
\begin{equation}\label{eq:o_msto}
\mathbf{o}(n)=\mathcal{T}_{\mathbf{H}(n),\lambda_2}(\mathbf{g}(n))=
\left\{\begin{array}{cc}
-(\mathbf{H}(n)+\gamma\mathbf{I}_p)^{-1}\mathbf{g}(n),&\textrm{if 
}\|\mathbf{g}(n)\|_2>\lambda_2\\
\mathbf{0}_p,& \textrm{otherwise}\end{array}\right.
\end{equation}
where parameter $\gamma:=\lambda_2^2/(2\eta)$ is such that $\eta>0$ solves the 
scalar optimization
\begin{equation}\label{eq:line_search}
\min_{\eta>0}\left(1-\mathbf{g}^\prime(n)\left(2\eta\mathbf{H}(n)+
\lambda_2^2\right)^{-1}\mathbf{g}(n)\right)\eta.
\end{equation}
Remarkably, one can easily determine if $\mathbf{o}(n)=\mathbf{0}_p$, 
by forming $\mathbf{g}(n)$ and checking whether $\|\mathbf{g}(n)\|_2\leq 
\lambda_2$. This will be the computational burden incurred to solve 
\eqref{eq:sparse_coding_o} for most $n$, since outliers are typically sporadic 
and one would expect to obtain $\mathbf{o}(n)=\mathbf{0}_p$ most of the time. 
When datum $\mathbf{x}_n$ is deemed an outlier, 
$\|\mathbf{g}(n)\|_2>\lambda_2$, and one needs to 
carry out the extra line search in \eqref{eq:line_search} to determine 
$\mathbf{o}(n)$
as per \eqref{eq:o_msto}; further details can be found in
in~\cite{puig_hero}. Whenever an $\ell_1$-norm outlier regularization is adopted,
the resulting counterpart of \eqref{eq:sparse_coding_o} can be solved using e.g.,
coordinate descent~\cite{da_jn_gg_2010}, or, the 
Lasso variant of least-angle regression (LARS)~\cite{mairal10}. 
 
Moving on, the subspace update is given by
\begin{equation*}\label{eq:C_update_online}
\mathbf{U}(n)=\arg\min_{\mathbf{U}}
\sum_{i=1}^{n}\beta^{n-i}\|\mathbf{x}_{i}-\mathbf{m}(i-1)-
\mathbf{U}\mathbf{s}(i)-\mathbf{o}(i)\|_2^2
\end{equation*}
and can be efficiently obtained from $\mathbf{U}(n-1)$, via a
recursive LS update leveraging the matrix inversion
lemma; see e.g.,~\cite{yang95}. Note that the orthonormality constraint on 
$\mathbf{U}$ is not enforced here, yet the deviation from
orthonormality is typically small as observed in~\cite{yang95}.
Still, if orthonormal principal directions are required, 
an extra orthonormalization step can be carried out per
iteration, or, once at the end of the process. Finally,
$\mathbf{m}(n)$ is obtained recursively as the
exponentially-weighted average of the outlier-compensated data
$\{\mathbf{x}_i-\mathbf{o}(i)\}_{i=1}^n$. The resulting online robust (OR-)PCA
algorithm and its initialization are summarized under Algorithm
\ref{table: AM_solver_online}, where $\mathbf{m}$ and its
update have been omitted for brevity.

\begin{algorithm}[t]
\caption{: Online robust (OR-)PCA} \small{
\begin{algorithmic}
    \STATE \begin{verbatim}\* Batch initialization phase\end{verbatim}
    \STATE Determine $\lambda_2$ and $\mathbf{U}(n_0)$ from $\{\mathbf{x}_n\}_{n=1}^{n_0}$, 
    as in Section \ref{ssec:params}. Initialize 
$\mathbf{P}(n_0)=10^3\mathbf{I}_p$ and $\mathbf{s}(n_0)=\mathbf{0}_q$.
    \STATE \begin{verbatim}\* Online phase\end{verbatim}
    \FOR {$n=n_0+1,n_0+2,\ldots$}
    	\STATE Form $\mathbf{H}(n)$ and $\mathbf{g}(n)$ using \eqref{eq:H} and \eqref{eq:g}.
        \STATE Update $\mathbf{o}(n)=\mathcal{T}_{\mathbf{H}(n),\lambda_2}(\mathbf{g}(n))$ via \eqref{eq:o_msto}.
        \STATE Update $\mathbf{s}(n)=\mathbf{U}^\prime(n-1)[\mathbf{x}_n-\mathbf{o}(n)]$.
        \STATE \begin{verbatim}\* RLS subspace update\end{verbatim}
        \STATE Update $\mathbf{k}(n)=\mathbf{P}(n-1)\mathbf{s}(n)/[\beta+\mathbf{s}^\prime(n)\mathbf{P}(n-1)\mathbf{s}(n)]$.	
	  \STATE Update $\mathbf{P}(n)=(1/\beta)[\mathbf{P}(n-1)-\mathbf{k}(n)(\mathbf{P}(n-1)\mathbf{s}(n))^\prime]$.	
        \STATE Update $\mathbf{U}(n)=\mathbf{U}(n-1)+[\mathbf{x}_n-\mathbf{U}(n-1)\mathbf{s}(n)-\mathbf{o}(n)]\mathbf{k}^\prime(n).$
    \ENDFOR
	\end{algorithmic}} \label{table: AM_solver_online}
\end{algorithm}

For the batch case where all data in $\mathcal{T}_x$ are available for joint 
processing, two data-driven criteria to select $\lambda_2$ have been outlined 
in Section \ref{ssec:params}. However, none of these 
sparsity-controlling mechanisms can be run in real-time, and selecting
$\lambda_2$ for subspace tracking via OR-PCA is challenging. 
One possibility to circumvent this problem is to select $\lambda_2$ once 
during a short initialization (batch) phase of OR-PCA, 
and retain its value for the subsequent time instants. 
Specifically, the initialization phase of OR-PCA entails 
solving \eqref{eq:cost_l1unconstr} using Algorithm \ref{table: AM_solver},
with a typically small batch of data $\{\mathbf{x}_n\}_{n=1}^{n_0}$. At time $n_0$, 
the criteria in Section \ref{ssec:params} 
are adopted to find the `best' $\lambda_2$, and thus obtain the subspace estimate
$\hat{\mathbf{U}}(n_0)$ required to initialize the OR-PCA iterations. 

Convergence analysis of OR-PCA algorithm
is beyond the scope of the present paper, and is only confirmed via 
simulations. The numerical tests in Section
\ref{sec:sims} also show that in the presence of outliers, the
novel adaptive algorithm outperforms existing non-robust
alternatives for subspace tracking.

% % % % % % % % % % % % % % % % % % % % % % % % % % % % % % % % % % % % % % % %
%                         Section VI                                          %
% % % % % % % % % % % % % % % % % % % % % % % % % % % % % % % % % % % % % % % %

\section{Robustifying Kernel PCA}\label{sec:kernel_PCA}

Kernel (K)PCA is a generalization to (linear) PCA, seeking
principal components in a \textit{feature space} nonlinearly
related to the \textit{input space} where the data in
$\mathcal{T}_x$ live~\cite{ssm97}. KPCA has been shown effective
in performing nonlinear feature extraction for pattern
recognition~\cite{ssm97}. In addition, connections between KPCA
and spectral clustering~\cite[p. 548]{HTF09} motivate well the novel
KPCA method developed in this section, to robustly identify
cohesive subgroups (communities) from social network data.

Consider a nonlinear function $\bm
\phi:\mathbb{R}^p\to\mathcal{H}$, that maps elements from the
input space $\mathbb{R}^p$ to a feature space $\mathcal{H}$ of
arbitrarily large -- possibly infinite -- dimensionality. Given transformed data
$\mathcal{T}_{\mathcal{H}}:=\{\bm\phi(\mathbf{x}_n)\}_{n=1}^{N}$, the
proposed approach to robust KPCA fits the model
\begin{equation}\label{eq:factor_model_fspace}
\bm\phi(\mathbf{x}_n)=\mathbf{m}+\mathbf{U}\mathbf{s}_n+\mathbf{e}_n+\mathbf{o}_n,\quad n=1,\ldots,N
\end{equation}
by solving ($\bm\Phi:=[\bm\phi(\mathbf{x}_1),\ldots,\bm\phi(\mathbf{x}_N)]$)
\begin{equation}\label{eq:frob_reg_kernel}
\min_{\mathbf{U},\mathbf{S},\mathbf{O}}
\|\bm\Phi^\prime-\mathbf{1}_{N}\mathbf{m}^\prime-\mathbf{S}\mathbf{U}^\prime-\mathbf{O}\|_F^2
+\frac{\lambda_{\ast}}{2}(\|\mathbf{U}\|_{F}^2+\|\mathbf{S}\|_{F}^2)+\lambda_2
\|\mathbf{O}\|_{2,r}.
\end{equation}
It is certainly possible to adopt the criterion \eqref{eq:cost_l1unconstr} as
well, but \eqref{eq:frob_reg_kernel} is
chosen here for simplicity in exposition. 
Except for the principal components' matrix $\mathbf{S}\in\mathbb{R}^{N\times \bar{q}}$, 
both the data and the unknowns in
\eqref{eq:frob_reg_kernel} are now vectors/matrices of generally infinite
dimension.  In 
principle, this 
challenges the optimization task
since it is impossible to store, or, perform updates of such
quantities directly. For these reasons, assuming zero-mean data 
$\bm\phi(\mathbf{x}_n)$, or, the possibility of mean compensation for that 
matter, cannot be taken for granted here [cf. Remark \ref{rem:mean}]. Thus, it 
is important to explicitly consider the estimation of  $\mathbf{m}$. 

Interestingly, this hurdle can be overcome by endowing
$\mathcal{H}$ with the structure of a reproducing kernel
Hilbert space (RKHS), where inner products between any two
members of $\mathcal{H}$ boil down to evaluations of the
reproducing kernel
$K_{\mathcal{H}}:\mathbb{R}^p\times\mathbb{R}^p \to\mathbb{R}$,
i.e.,
$\langle\bm\phi(\mathbf{x}_i),\bm\phi(\mathbf{x}_j)\rangle_\mathcal{H}=K_{\mathcal{H}}(\mathbf{x}_i,\mathbf{x}_j)$.
Specifically, it is possible to form the kernel matrix
$\mathbf{K}:=\bm\Phi^\prime\bm\Phi\in\mathbb{R}^{N\times N}$, without
directly working with the vectors in $\mathcal{H}$.
This so-termed \textit{kernel trick} is the crux of most kernel
methods in machine learning~\cite{HTF09}, including kernel
PCA~\cite{ssm97}. The problem of selecting 
$K_\mathcal{H}$ (and $\bm\phi$ indirectly) will not be
considered here.

Building on these ideas, it is shown in the sequel
that Algorithm \ref{table: AM_solver_rank_control} can be \textit{kernelized},
to solve \eqref{eq:frob_reg_kernel} at affordable computational complexity 
and memory storage requirements that 
do not depend on the dimensionality of $\mathcal{H}$.

\begin{proposition}\label{prop:kernelized}
For $k\geq 1$, the sequence of iterates generated
by Algorithm \ref{table: AM_solver_rank_control} when applied to solve
\eqref{eq:frob_reg_kernel} can be written as $\mathbf{m}(k)=\bm
\Phi\bm\mu(k)$, $\mathbf{U}(k)=\bm\Phi\bm\Upsilon(k)$, and 
$\mathbf{O}^\prime(k)=\bm\Phi\bm\Omega(k)$. The quantities 
$\bm\mu(k)\in\mathbb{R}^N$, $\bm\Upsilon(k)\in\mathbb{R}^{N\times \bar q}$, and 
$\bm\Omega(k)\in\mathbb{R}^{N\times N}$ are recursively updated as in 
Algorithm \ref{table: AM_solver_kernel}, without the need of operating with 
vectors in $\mathcal{H}$.
\end{proposition}
\begin{IEEEproof}
The proof relies on an inductive argument. Suppose that at iteration $k-1$, 
there exists a matrix $\bm{\Omega}(k-1)\in\mathbb{R}^{N\times N}$ such that 
the outliers can be expressed as 
$\mathbf{O}^\prime(k-1)=\bm\Phi\bm\Omega(k-1)$.  
From Algorithm \ref{table: AM_solver_rank_control}, the update for 
the mean vector is
$\mathbf{m}(k)=[\bm\Phi^\prime-\mathbf{O}(k-1)]^\prime\mathbf{1}_N/N=[\bm\Phi-
\bm\Phi\bm\Omega(k-1)]\mathbf{1}_N/N=\bm\Phi\bm\mu(k)$
where $\bm\mu(k):=[\mathbf{I}_n-\bm\Omega(k-1)]\mathbf{1}_N/N$. Likewise, 
$\mathbf{X}_o(k)=\bm\Phi^\prime-\mathbf{1}_N\bm\mu^\prime(k)\bm\Phi^\prime-\bm
\Omega^\prime(k-1)\bm\Phi^\prime$ so that one can write the subspace update as 
$\mathbf{U}(k)=\bm\Phi\bm\Upsilon(k)$, upon defining
\begin{equation*}
\bm\Upsilon(k):=[\mathbf{I}_N-\bm\mu(k)\mathbf{1}_N^{\prime}-\bm 
\Omega(k-1)]\mathbf{S}(k-1)[\mathbf{S}^\prime(k-1)\mathbf{S}(k-1)+(\lambda_\ast
/2)\mathbf{I}_{\bar{q}}]^{-1}.
\end{equation*}
With regards to the principal components, it follows that (cf. Algorithm \ref{table: AM_solver_rank_control})
\begin{align}
\mathbf{S}(k)={}&[\mathbf{I}_N-\mathbf{1}_N\bm\mu^\prime(k)-\bm\Omega^\prime(k-
1)]
\bm\Phi^\prime\bm\Phi\bm\Upsilon(k)[\bm\Upsilon(k)^\prime\bm\Phi^\prime
\bm\Phi\bm\Upsilon(k)+(\lambda_\ast/2)\mathbf{I}_{\bar{q}}]^{-1}\nonumber\\
={}&[\mathbf{I}_N-\mathbf{1}_N\bm\mu^\prime(k)-\bm\Omega^\prime(k-1)]
\mathbf{K}\bm\Upsilon(k)[\bm\Upsilon(k)^\prime\mathbf{K}\bm\Upsilon(k)+(\lambda
_\ast/2)\mathbf{I}_{\bar{q}}]^{-1}\label{eq:S_kernelized}
\end{align}
which is expressible in terms of the kernel matrix 
$\mathbf{K}:=\bm\Phi^\prime\bm\Phi$. Finally, the columns $\mathbf{o}_n(k)$ 
are given by the vector soft-thresholding operation \eqref{eq:vector_soft_t}, 
where the residuals are
\begin{equation*}
\mathbf{r}_n(k)=\bm\phi(\mathbf{x}_n)-\mathbf{m}(k)-\mathbf{U}(k)\mathbf{s}_n(k
)=
\bm{\Phi}[\mathbf{b}_{N,n}-\bm\mu(k)-\bm{\Upsilon(k)}\mathbf{s}_n(k)]:=\bm{\Phi
}\bm{\rho}_n(k).
\end{equation*}
Upon stacking all columns $\mathbf{o}_n(k),$ $n=1,\ldots,N$, one readily 
obtains [cf. \eqref{eq:vector_soft_t}]
\begin{equation}\label{eq:O_prime}
\mathbf{O}^\prime(k)=
\bm{\Phi}[\mathbf{I}_{N}-\bm\mu(k)\mathbf{1}_N^\prime-\bm{\Upsilon(k)}
\mathbf{S}^\prime(k)]\bm{\Lambda}(k)
\end{equation}
where 
$\bm{\Lambda}(k):=\textrm{diag}((\|\mathbf{r}_1(k)\|_2-\lambda_2/2)_+/\|\mathbf
{r}_1(k)\|_2
,\ldots,(\|\mathbf{r}_N(k)\|_2-\lambda_2/2)_+/\|\mathbf{r}_N(k)\|_2)$. 
Interestingly, the diagonal elements of $\bm{\Lambda}(k)$ can be computed 
using the kernel matrix, since 
$\|\mathbf{r}_n(k)\|_2=\sqrt{\bm{\rho}_n^\prime(k)\mathbf{K}\bm{\rho}_n(k)}$, 
$n=1,\ldots,N$. From \eqref{eq:O_prime} it is apparent that one can write 
$\mathbf{O}^\prime(k)=\bm{\Phi}\bm{\Omega}(k)$, after defining
\begin{equation*}
\bm{\Omega}(k):=[\mathbf{I}_{N}-\bm\mu(k)\mathbf{1}_N^\prime-\bm{\Upsilon(k)}
\mathbf{S}^\prime(k)]\bm{\Lambda}(k).
\end{equation*}
The proof is concluded by noting that for $k=0$, Algorithm \ref{table: 
AM_solver_rank_control} is initialized with 
$\mathbf{O}^\prime(0)=\mathbf{0}_{p\times N}$. One can thus satisfy the 
inductive base case $\mathbf{O}^\prime(0)=\bm{\Phi}\bm{\Omega}(0)$, by letting 
$\bm{\Omega}(0)=\mathbf{0}_{N\times N}$.
\end{IEEEproof}

In order to run the novel robust KPCA algorithm (tabulated as Algorithm 
\ref{table: AM_solver_kernel}), one does not have to 
store or process the quantities $\mathbf{m}(k)$, $\mathbf{U}(k)$, and 
$\bbO(k)$. As per Proposition \ref{prop:kernelized}, the iterations of the
provably convergent AM solver in Section \ref{ssec:Frob} can be equivalently 
carried out by cycling through \textit{finite-dimensional} `sufficient 
statistics'
$\bm\mu(k)\to\bm\Upsilon(k)\to\mathbf{S}(k)\to\bm\Omega(k)$. In other words, 
the iterations of the robust kernel PCA algorithm are devoid of algebraic 
operations among vectors in $\mathcal{H}$. Recall that the 
size of matrix $\mathbf{S}$ is independent of the dimensionality of 
$\mathcal{H}$. 
Nevertheless, its update in Algorithm \ref{table: AM_solver_rank_control} 
cannot be carried out verbatim in the high-dimensional setting here, and is 
instead kernelized to yield the update rule \eqref{eq:S_kernelized}. 

Because 
$\mathbf{O}^\prime(k)=\bm\Phi\bm\Omega(k)$ and upon convergence of the 
algorithm, the outlier vector
norms are computable in terms of $\mathbf{K}$, i.e.,
$[\|\mathbf{o}_1(\infty)\|_2^2,\ldots,\|\mathbf{o}_N(\infty)\|_2^2]^\prime=
\textrm{diag}[\bm\Omega^\prime(\infty)\mathbf{K}\bm\Omega(\infty)]$. These are 
critical to determine the robustification paths needed to carry out the 
outlier sparsity control methods in Section \ref{ssec:params}.
Moreover, the principal component corresponding to any given new data point 
$\mathbf{x}$ is obtained through the projection
$\mathbf{s}=\mathbf{U}(\infty)^\prime[\bm\phi(\mathbf{x})-\mathbf{m}(\infty)]
=\bm\Upsilon^\prime(\infty)\bm\Phi^\prime\bm\phi(\mathbf{x})-
\bm\Upsilon^\prime(\infty)\mathbf{K}\bm\mu(\infty)$,
which is again computable after $N$ evaluations the kernel function
$K_\mathcal{H}$.

\begin{algorithm}[t]
\caption{: Robust KPCA solver} \small{
\begin{algorithmic}
    \STATE Initialize $\bm\Omega(0)=\mathbf{0}_{N\times N}$, $\mathbf{S}(0)$ 
randomly, and form $\mathbf{K}=\bm\Phi^\prime\bm\Phi$.
    \FOR {$k=1,2,\ldots$}
        \STATE Update $\bm\mu(k)=[\mathbf{I}_n-\bm\Omega(k-1)]\mathbf{1}_N/N.$
        \STATE Form
        $\bm\Phi_o(k)=\mathbf{I}_N-\bm\mu(k)\mathbf{1}_N^{\prime}-\bm 
        \Omega(k-1).$
        \STATE Update 
$\bm\Upsilon(k)=\bm\Phi_o(k)\mathbf{S}(k-1)[\mathbf{S}^\prime(k-1)\mathbf{S}(k-
1)+(\lambda_\ast
        /2)\mathbf{I}_{\bar{q}}]^{-1}.$
	  \STATE Update
$\mathbf{S}(k)=\bm\Phi_o^\prime(k)\mathbf{K}\bm\Upsilon(k)
[\bm\Upsilon(k)^\prime\mathbf{K}\bm\Upsilon(k)+(\lambda_\ast/2)\mathbf{I}_{\bar
{q}}]^{-1}.$
	\STATE Form 
$\bm\rho_n(k)=\mathbf{b}_{N,n}-\bm\mu(k)-\bm{\Upsilon(k)}\mathbf{s}_n(k)$, 
$n=1,\ldots,N$, and update $\bm\Lambda(k)$. 
	\STATE Update 
$\bm{\Omega}(k)=[\mathbf{I}_{N}-\bm\mu(k)\mathbf{1}_N^\prime-\bm{\Upsilon(k
)}
	\mathbf{S}^\prime(k)]\bm{\Lambda}(k).$
    \ENDFOR
	\end{algorithmic}} \label{table: AM_solver_kernel}
\end{algorithm}

% % % % % % % % % % % % % % % % % % % % % % % % % % % % % % % % % % % % % % % %
%                         Section VII                                         %
% % % % % % % % % % % % % % % % % % % % % % % % % % % % % % % % % % % % % % % %

\vspace{-0.1cm}
\section{Numerical Tests}\label{sec:sims}

\subsection{Synthetic data tests}
\label{ssec:synthetic}

To corroborate the effectiveness of the proposed robust methods, 
experiments with computer generated data are carried out first. These
are important since they provide a `ground truth', against which performance
can be assessed by evaluating suitable figures of merit.
 
\noindent\textbf{Outlier-sparsity control.} To generate the data
\eqref{eq:model_outliers}, a similar setting as in~\cite[Sec. V]{zlwcm10}
is considered here with $N=p$ and $\mathbf{m}=\mathbf{0}_p$. 
For $n=1,\ldots,N$, the errors are $\mathbf{e}_n\sim
\mathcal{N}(\mathbf{0}_p,\sigma_e^2\mathbf{I}_p)$ (multivariate normal distribution) and i.i.d. 
The entries of $\mathbf{U}$ and $\{\mathbf{s}_n\}_{n=1}^N$ are i.i.d. zero-mean
Gaussian distributed, with variance $\sigma_{U,s}^2=10\sigma_e/\sqrt{N}$. Outliers
are generated as $\mathbf{o}_n = \mathbf{p}_n \odot \mathbf{q}_n$, where 
the entries of $\mathbf{p}_n$ are i.i.d. Bernoulli distributed with parameter $\rho_p$,
and $\mathbf{q}_n$ has i.i.d. entries drawn from a uniform distribution supported
on $[-5,5]$. The chosen values of the 
parameters are $N=p=200$, $q=20$, $\rho_p=0.01$, and varying noise levels 
$\sigma_e^2=\{0.01,0.05,0.1,0.25,0.5\}$.

In this setup, the ability to recover the low-rank component of the data 
$\mathbf{L}:=\mathbf{S}\mathbf{U}'$ is tested for the sparsity-controlling
robust PCA method of this paper [cf. \eqref{eq:cost_l1unconstr}], 
stable PCP \eqref{eq:spcp}, 
and (non-robust) PCA. The $\ell_1$-norm regularized counterparts of \eqref{eq:cost_l1unconstr}
and \eqref{eq:spcp} are adopted to deal with entry-wise outliers. Both 
values of $q$ and $\sigma_e^2$ are assumed known to obtain
$\hat{\mathbf{L}}:=\hat{\mathbf{S}}\hat{\mathbf{U}}'$ and $\hat{\mathbf{O}}$ 
via 
\eqref{eq:cost_l1unconstr}. This way, $\lambda_2$ is chosen using the sparsity-controlling
algorithm of Section \ref{ssec:params}, searching over a grid where $G_{\lambda}=200$,  
$\lambda_{\min}=10^{-2}\lambda_{\max}$, and $\lambda_{\max}=20$. In addition, the
solutions of \eqref{eq:cost_l1unconstr} are refined by running two iterations of the iteratively 
reweighted algorithm in Section \ref{ssec:nonconvex}, where $\delta=10^{-5}$. 
Regarding SPCP, only the knowledge of 
$\sigma_e^2$ is required to select the tuning parameters 
$\lambda_\ast=2\sqrt{2N\sigma_e^2}$ and $\lambda_2=2\sqrt{2\sigma_e^2}$ in \eqref{eq:spcp},
 as suggested in~\cite{zlwcm10}. Finally, the best rank $q$ approximation to
 the data $\mathbf{X}$ is obtained using standard PCA.
 
The results are summarized in Table \ref{table:results}, which shows the
estimation errors $\bar{\textrm{err}}:=\|\mathbf{L}-\hat{\mathbf{L}}\|_F/N$
 attained by the aforementioned schemes,  averaged over $15$ runs of the experiment.
The `best' tuning parameters $\lambda_2^\ast$ used in \eqref{eq:cost_l1unconstr} are also shown. 
Both robust schemes attain an error which is approximately an order of magnitude
smaller than PCA. With the additional knowledge of the true data rank $q$, the 
sparsity-controlling algorithm of this paper outperforms stable PCP in 
terms of $\bar{\textrm{err}}$. This numerical test is used to validate 
Proposition \ref{prop:prop_equiv_frob} 
as well. For the same values of the tuning parameters chosen for 
\eqref{eq:spcp} and the rank upper-bound set to $\bar{q}=2q$, Algorithm 
\ref{table: 
AM_solver_rank_control} is run to obtain the solution 
$\{\bar{\mathbf{U}},\bar{\mathbf{S}},\bar{\mathbf{O}}\}$ of the nonconvex 
problem \eqref{eq:frob_reg}. The average (across realizations and values of 
$\sigma_e^2$) errors obtained are  
$\|\hat{\mathbf{L}}-\bar{\mathbf{S}}\bar{\mathbf{U}}^\prime\|_F/N=0.15\times 
10^{-6}$ and 
$\|\hat{\mathbf{O}}-\bar{\mathbf{O}}\|_F/N=0.78\times 
10^{-7}$, where 
$\{\hat{\mathbf{L}},\hat{\mathbf{O}}\}$ is the solution of stable PCP [cf. 
\eqref{eq:spcp}]. Thus, the solutions are identical for all practical 
purposes.
 
\noindent\textbf{Identification of invalid survey protocols.} 
Robust PCA is tested here to identify invalid
or otherwise aberrant item response (questionnaire) data in surveys, that is, 
to flag and hold in abeyance data that may
negatively influence (i.e., bias) subsequent data summaries and
statistical analyses. In recent years, item response theory (IRT)
has become the dominant paradigm for constructing and
evaluating questionnaires in the biobehavioral and health
sciences and in high-stakes testing (e.g., in the development
of college admission tests); see e.g.,~\cite{wr10chap}. 
IRT entails a class of nonlinear models characterizing an
individual's item response behavior by one or more latent
traits, and one or more item parameters. An increasingly popular IRT model
for survey data is the 2-parameter logistic IRT model
(2PLM)~\cite{rw93jpsp}.  2PLM characterizes  the probability of
a keyed (endorsed) response $y_{nm}$, as a nonlinear function
of a weighted difference between a person
parameter $\theta_n$ and an item parameter $b_m$%.  The
%equation for a 2PLM item response function is shown below.
%
\begin{equation}
{\rm Pr}(y_{nm}=1|\theta_n)= 
\frac{e^{1.7a_m(\theta_n-b_m)}}{1+e^{1.7a_m(\theta_n-b_m)}}
\label{2PLM}
\end{equation}
where $\theta_n$ is a latent trait value for individual $n$;
$a_m$ is an item discrimination parameter (similar to a factor
loading) for item $m$; and $b_m$ is an item difficulty (or
extremity) parameter for item $m$. 
%One reason for the popularity of 2PLM is that under certain assumptions, its
%parameters can be transformed into the person and item parameters of the 
%maximum likelihood factor analysis model~\cite{td87}.

Binary item responses (`agree/disagree' response
format) were generated for $N=1,000$ hypothetical subjects who were 
administered
$p=200$ items (questions).  The 2PLM function \eqref{2PLM}
was used to generate the underlying item response
probabilities, which were converted into binary item responses
as follows: a response was coded 1 if
$\textrm{Pr}(y_{nm}|\theta_n) \ge \mathcal{U}(0,1)$, and coded
0 otherwise, where $\mathcal{U}[0,1]$ denotes a uniform random
deviate over $[0,1]$. Model parameters were randomly
drawn as $\{a_m\}_{m=1}^{200}\sim \mathcal{U}[1, 1.5]$,
$\{b_m\}_{m=1}^{200}\sim \mathcal{U}[-2, 2]$, and
$\{\bm\theta_l\}_{l=1}^{200}\sim
\mathcal{N}(\mathbf{0}_5,\mathbf{I}_{5}$). % were randomly sampled from a
%$N(\textbf{0},\textbf{I})$ distribution, where $\textbf{I}$ is
%a 5 $\times$ 5 identity matrix.
Each of the 200 items loaded on one of $q=5$ latent factors. To simulate 
random responding -- a prevalent 
form of aberrancy
in e.g., web-collected data -- rows 101-120 of the
item response matrix $\mathbf{Y}$ were modified by (re)drawing each of the 
entries from a
Bernoulli distribution with parameter 0.5, thus yielding the corrupted matrix 
$\mathbf{X}$. 

\begin{figure}[h]
\begin{center}
\includegraphics[width=0.45\linewidth]{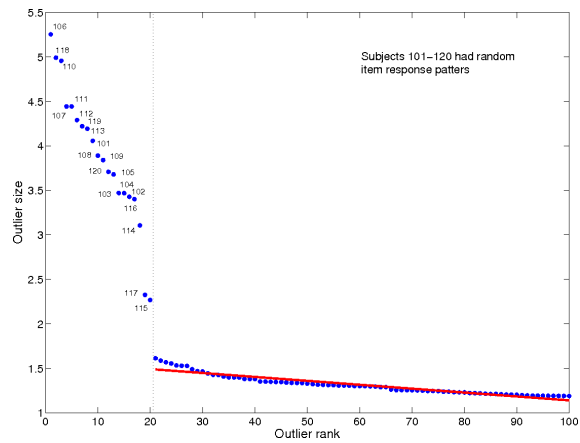}\\
%\vspace{-0.5cm}
\caption{Pseudo scree plot of outlier size
$(\|\hat{\mathbf{o}}_n\|_2)$; the 100 largest outliers are 
shown.}\label{fig:Fig_1}
\end{center}
\end{figure}

Robust PCA in 
\eqref{eq:cost_l1unconstr} was adopted to identify invalid 
survey data, with $q=5$, and $\lambda_2$
chosen such that $\|\hat{\mathbf{O}}\|_0=150$, 
a safe overestimate of the number of outliers. 
Results of this study are summarized in Fig.
\ref{fig:Fig_1}, which
displays the 100 largest outliers ($\|\hat{\mathbf{o}}_n\|_2$) 
from the robust PCA analysis of the
$N=1,000$ simulated response vectors. When the outliers are plotted
against their ranks, there is an unmistakable  break between
the $20$th and $21$st ordered value indicating that the
method correctly identified the \textit{number} of aberrant
response patterns in $\mathbf{X}$. Perhaps more impressively, the
method also correctly identified rows 101-to-120 as containing
the invalid data.

\begin{figure}[h]
\begin{minipage}[b]{0.5\linewidth}
  \centering
  \centerline{\includegraphics[width=\linewidth]{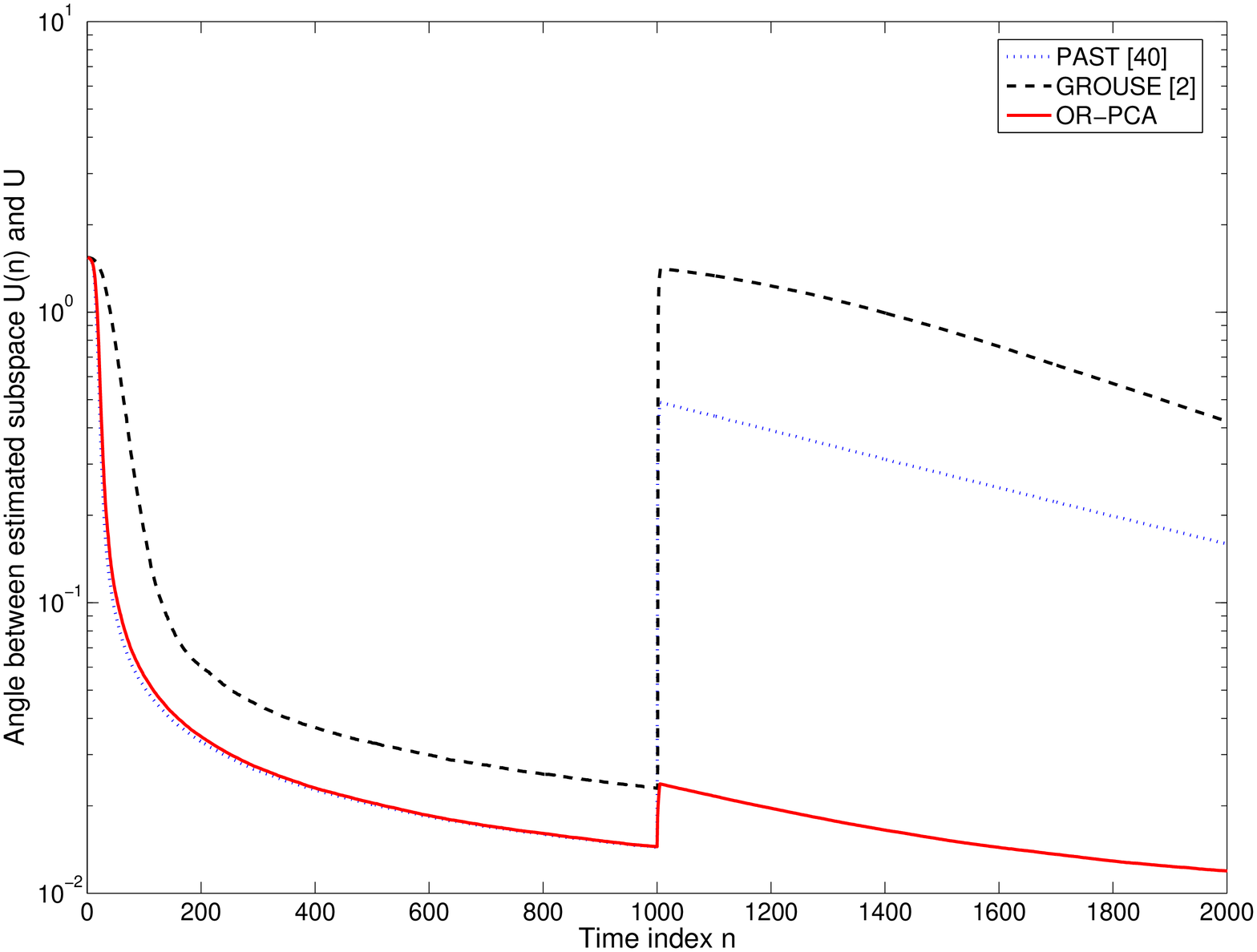}}
\medskip
\end{minipage}
\hfill
\begin{minipage}[b]{.5\linewidth}
  \centering
  \centerline{\includegraphics[width=\linewidth]{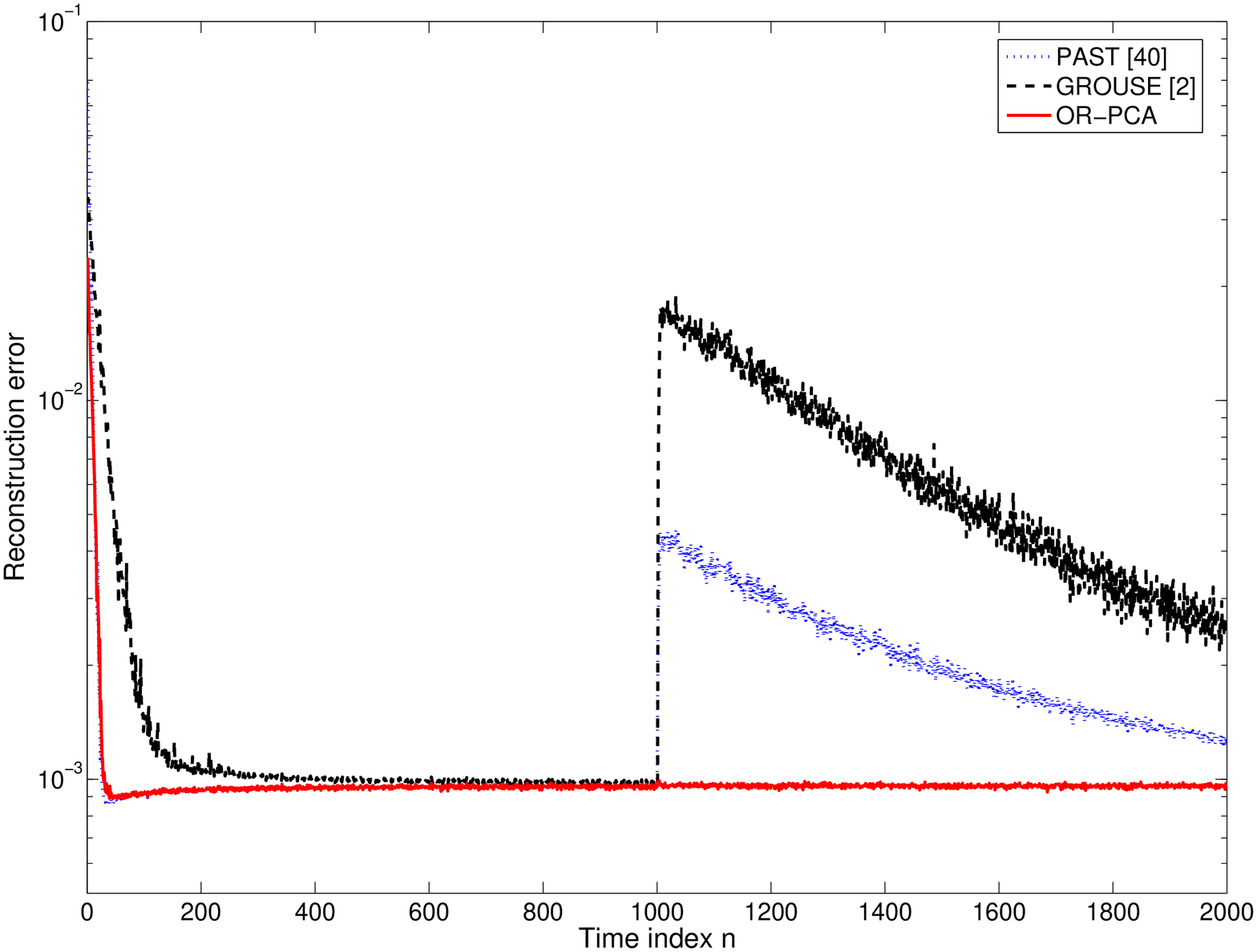}}
\medskip
\end{minipage}
%\vspace{-1cm}
\caption{(Left) Time evolution of the angle
between the learnt subspace $\mathbf{U}(n)$, and the true
$\mathbf{U}$ used to generate the data ($\beta=0.99$ and $\lambda_2=1.65$). 
Outlier contaminated data is introduced at time
$n=1001$. (Right) Time evolution of the reconstruction error.}
\label{fig:Fig_2} %\vspace{-0.7cm}
\end{figure}

\noindent\textbf{Online robust subspace estimation.} A simulated test is
carried out here to corroborate the convergence and
effectiveness of the OR-PCA algorithm in Section
\ref{sec:online}. For $N=2,000$, $p=150$, and $q=5$, nominal
data in $\mathcal{T}_y$ are generated according to the stationary model
\eqref{eq:factor_model}, where
$\mathbf{e}_n\sim\mathcal{N}(\mathbf{0}_p,10^{-3}\mathbf{I}_p)$.
Vectors $\mathbf{x}_{1001},\ldots,\mathbf{x}_{1005}$ are outliers, 
uniformly i.i.d. over $[-0.5,0.5]$. The results depicted in Fig. 
\ref{fig:Fig_2} are obtained after averaging over $50$ runs. Fig. 
\ref{fig:Fig_2} (left) depicts the time evolution of the 
angle
between the learnt subspace (spanned by the columns of)
$\hat{\mathbf{U}}(n)$ and the true subspace $\mathbf{U}$ generating
$\mathcal{T}_y$, where $\lambda_2=1.65$ and $\beta=0.99$. 
The convergent trend of Algorithm \ref{table: AM_solver_online} 
to $\mathbf{U}$ is apparent; and markedly
outperforms the non-robust subspace tracking method
in~\cite{yang95}, and the first-order GROUSE 
algorithm in~\cite{balzano_tracking}. Note that even though $\mathbf{U}$ 
is time-invariant, it is meaningful to select $0\ll\beta<1$ to quickly
`forget' and recover from the outliers. A similar trend can be observed
in Fig. \ref{fig:Fig_2} (right), which depicts the time evolution of
the reconstruction error 
$\|\mathbf{y}_n-\hat{\mathbf{U}}(n)\hat{\mathbf{U}}(n)^\prime\mathbf{y}_n\|_2^2/p$.

\begin{figure}[h]
\begin{minipage}[b]{0.5\linewidth}
  \centering
  \centerline{\includegraphics[width=\linewidth]{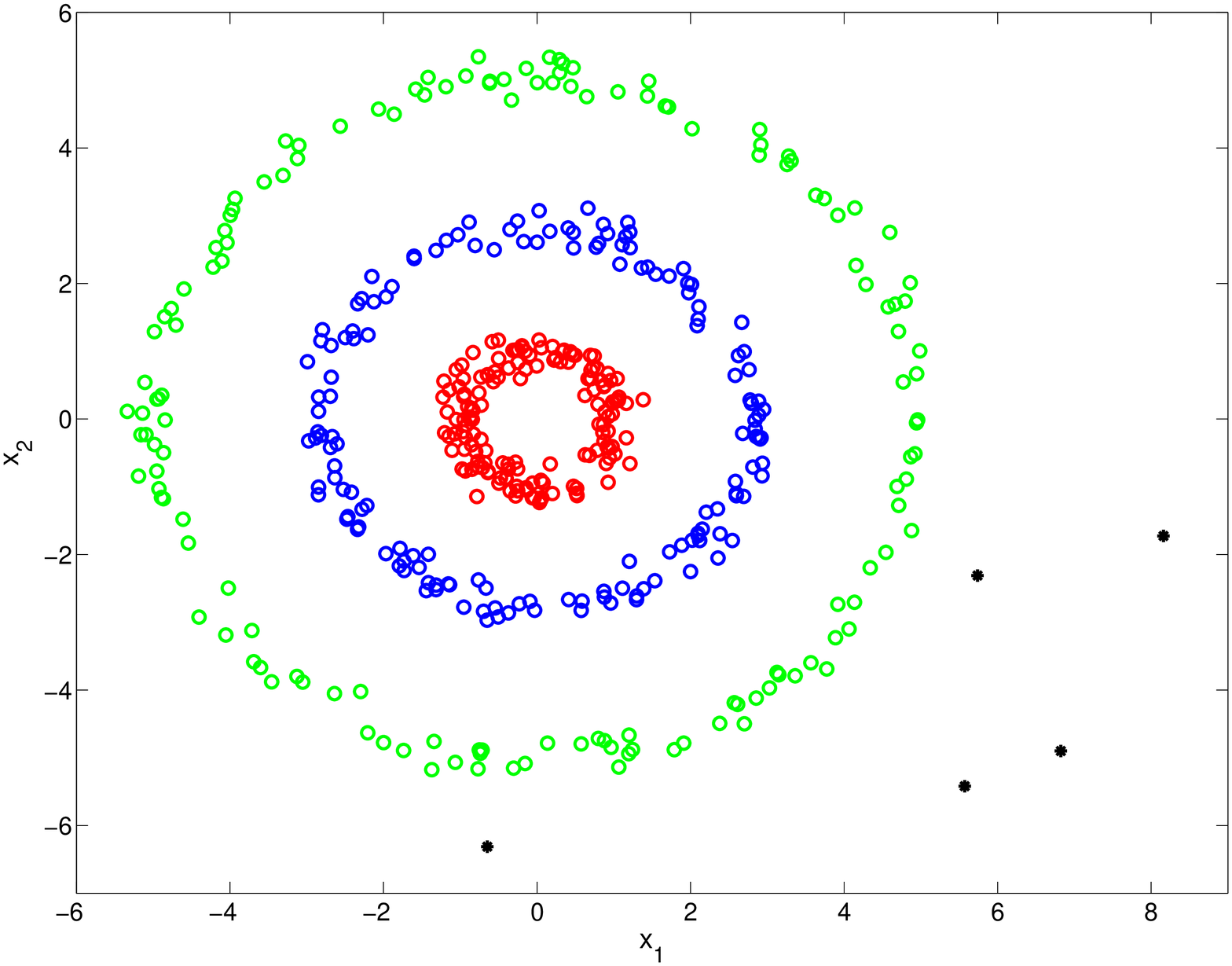}}
\medskip
\end{minipage}
\hfill
\begin{minipage}[b]{.5\linewidth}
  \centering
  \centerline{\includegraphics[width=\linewidth]{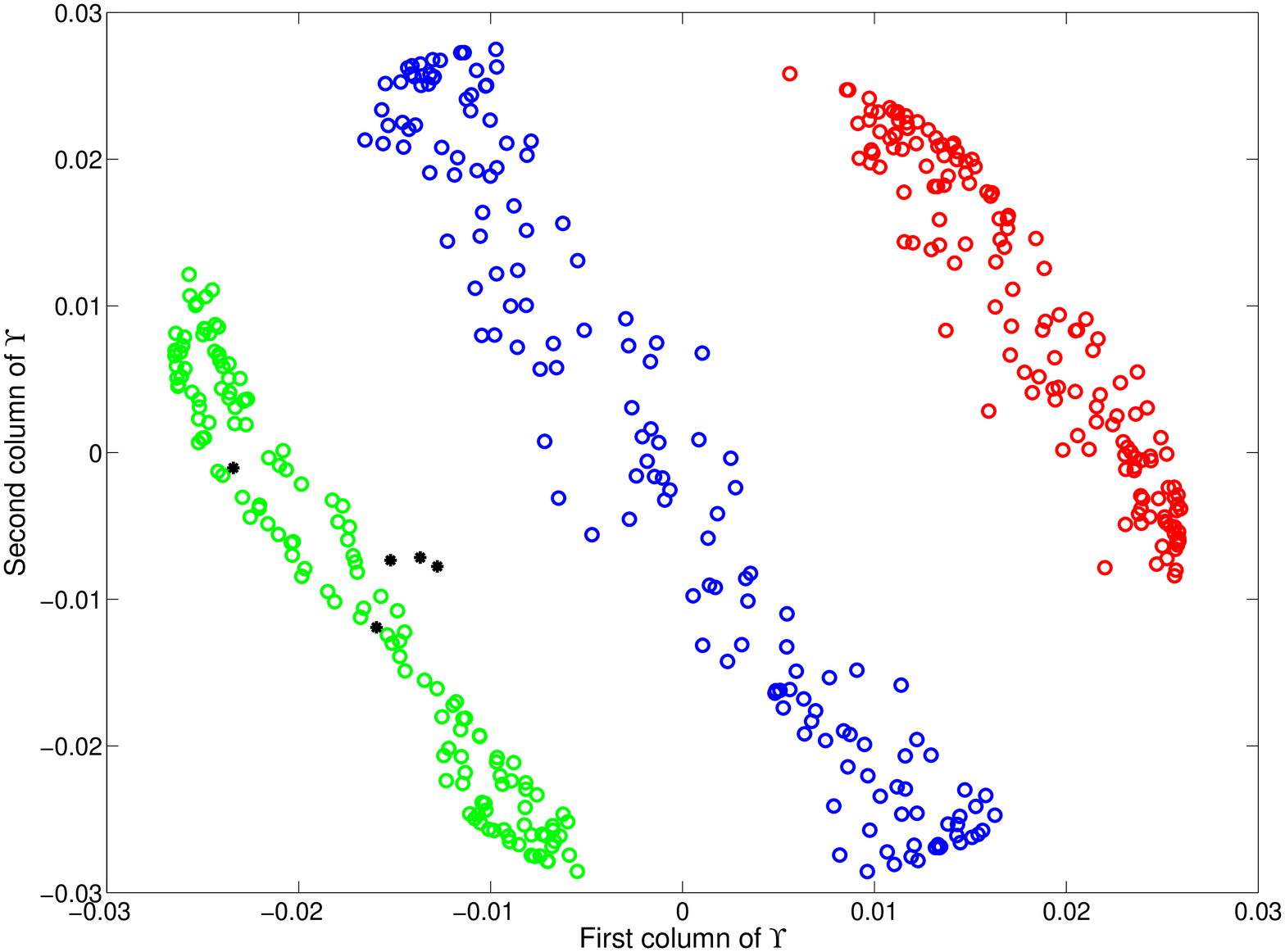}}
\medskip
\end{minipage}
%\vspace{-1cm}
\caption{(Left) Data in three concentric clusters, in addition to five 
outliers shown in black. 
(Right) Coordinates of the first two columns of $\bm{\Upsilon}$, obtained by running Algorithm 
\ref{table: AM_solver_kernel}. The five outlying points are correctly identified, and thus can be 
discarded. Non-robust methods will assign them to the green cluster.}
\label{fig:Fig_3} %\vspace{-1cm}
\end{figure}

\noindent\textbf{Robust spectral clustering.} The following simulated
test demonstrates that robust KPCA in Section \ref{sec:kernel_PCA}
can be effectively used to robustify spectral clustering (cf.
the connection between both non-robust methods in e.g.,~\cite[p. 548]{HTF09}). Adopting
the data setting from~\cite[p. 546]{HTF09}), $N=450$ points in $\mathbb{R}^2$ 
are generated from three circular concentric clusters, with respective radii of
$1$, $2.8$, and $5$. The points are uniformly distributed in angle, and
additive noise $\mathbf{e}_n\sim\mathcal{N}(\mathbf{0}_2,0.15\mathbf{I}_2)$
is added to each datum. Five outliers $\{\mathbf{x}_n\}_{n=451}^{455}$
uniformly distributed in the square $[-7,7]^2$ complete the training 
data $\mathcal{T}_x$; see Fig. \ref{fig:Fig_3} (left). To unveil
the cluster structure from the data, Algorithm \ref{table: AM_solver_kernel} is run
using the Gaussian radial kernel 
$K(\mathbf{x}_i,\mathbf{x}_j)=\textrm{exp}(-\|\mathbf{x}_i-\mathbf{x}_j\|_2^2/c)$,
with $c=10$. The sparsity-controlling parameter is set to $\lambda_2=1.85$ 
so that
$\|\hat{\mathbf{O}}\|_0=5$, while $\lambda_\ast=1$, and $\bar{q}=2$.
Upon convergence, the vector of estimated outlier norms is
$[\|\mathbf{o}_1(\infty)\|_2^2,\ldots,\|\mathbf{o}_{N+5}(\infty)\|_2^2]^\prime=[0,\ldots,0,
10^{-4},1.3\times 10^{-3},1.5\times 10^{-2}, 10^{-2},1.7\times 10^{-2}]^\prime$,
which shows that the outliers are correctly identified.
Estimates of the (rotated) first two dominant eigenvectors of the kernel matrix $\mathbf{K}$
are obtained as the columns of $\hat{\bm\Upsilon}$, and are depicted in 
Fig. \ref{fig:Fig_3} (right). After removing the rows of $\hat{\bm\Upsilon}$ 
corresponding to the outliers [black points in Fig. \ref{fig:Fig_3} (right)], e.g., K-means
clustering of the remaining points in Fig. \ref{fig:Fig_3} (right) will easily reveal the three 
clusters sought. From Fig. \ref{fig:Fig_3} (right)
it is apparent that a non-robust KPCA method will incorrectly 
assign the outliers to the outer (green) cluster.

\subsection{Real data tests}\label{ssec:real}

\noindent\textbf{Video surveillance.} To validate the proposed
approach to robust PCA, Algorithm \ref{table: AM_solver} was
tested to perform background modeling from a sequence of video
frames; an approach that has found widespread applicability for
intrusion detection in video surveillance systems. The
experiments were carried out using the dataset studied
in~\cite{dlt03}, which consists of $N=520$ images $(p=120\times
160)$ acquired from a static camera during two days. The
illumination changes considerably over the two day span, while
approximately $40\%$ of the training images contain people in
various locations. For $q=10$, both standard PCA and the robust
PCA of Section \ref{sec:spacor} were applied to build a
low-rank background model of the environment captured by the
camera. For robust PCA, $\ell_1$-norm regularization on
$\mathbf{O}$ was adopted to identify outliers at a pixel level.
The outlier sparsity-controlling parameter was chosen as
$\lambda_2=9.69\times 10^{-4}$, whereas a single iteration of
the reweighted scheme in Section \ref{ssec:nonconvex} was run
to reduce the bias in $\hat{\mathbf{O}}$.

Results are shown in Fig. \ref{fig:Fig_1}, for three
representative images. The first column comprises the original
frames from the training set, while the second column shows the
corresponding PCA image reconstructions. The
presence of undesirable `ghostly' artifacts is apparent, since
PCA is unable to completely separate the people from the
background. The third column illustrates the robust PCA
reconstructions, which recover the illumination changes while
successfully subtracting the people. The fourth column shows
the reshaped outlier vectors $\hat{\mathbf{o}}_n$, which mostly
capture the people and abrupt changes in illumination.

\noindent\textbf{Robust measurement of the Big Five personality 
factors.} The `Big Five' are five factors ($q=5$) of personality traits, 
namely extraversion, agreeableness, conscientiousness, neuroticism, and 
openness; see e.g.,~\cite{BF_chapter}. The Big Five inventory (BFI) on the 
other hand, is a brief 
questionnaire ($44$ items in total) tailored to measure the Big 
Five dimensions. Subjects taking the questionnaire are asked to 
rate in a scale from $1$ (disagree strongly) to $5$ (agree strongly), 
items of the form `I see myself as someone who is talkative'. Each item 
consists of a short phrase correlating (positively 
or negatively) with one factor; see e.g.,~\cite[pp. 157-58]{BF_chapter} for a 
copy of the BFI and scoring instructions.

\begin{figure}[h]
\begin{minipage}[b]{0.5\linewidth}
  \centering
  \centerline{\includegraphics[width=\linewidth]{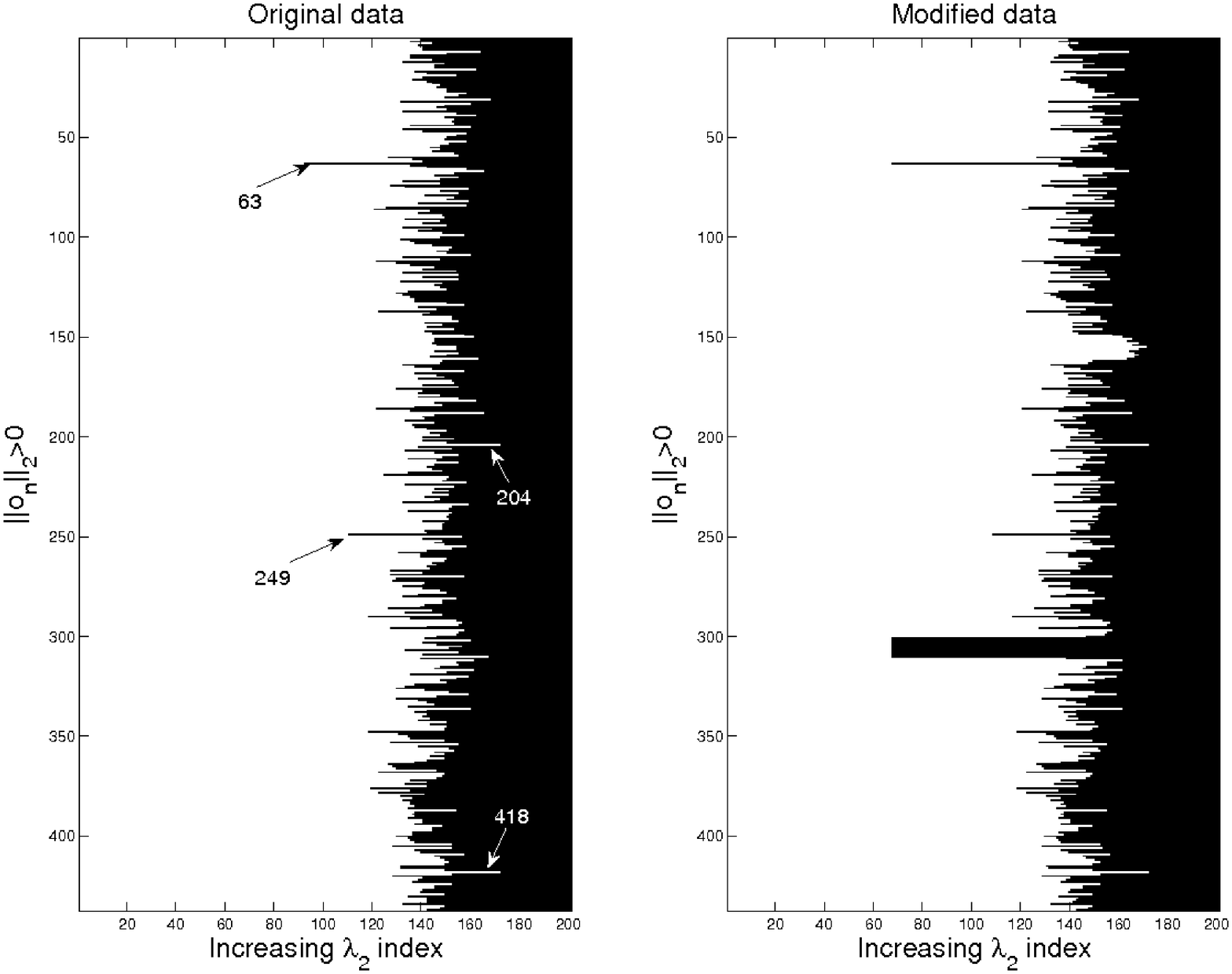}}
\medskip
\end{minipage}
\hfill
\begin{minipage}[b]{.5\linewidth}
  \centering
  \centerline{\includegraphics[width=\linewidth]{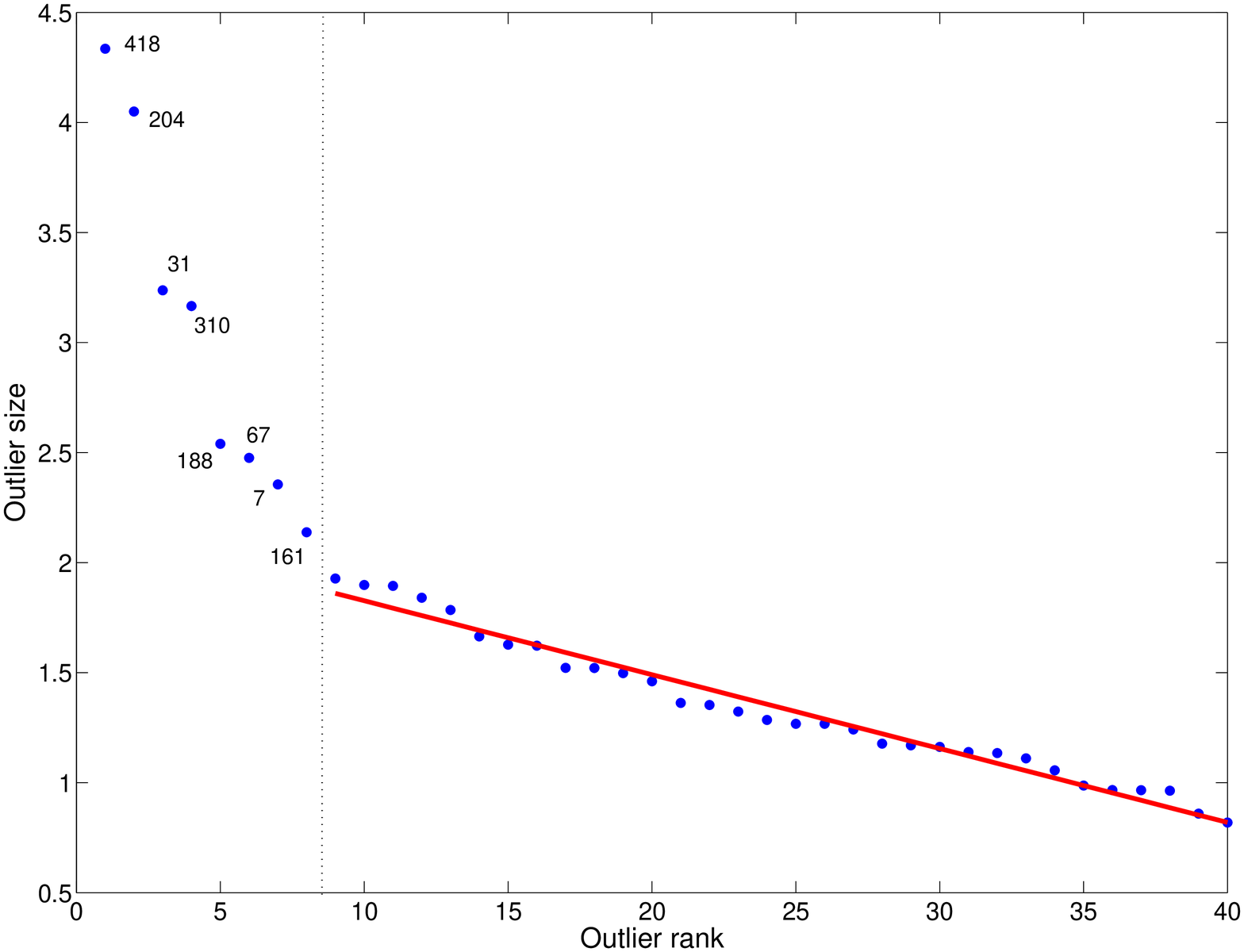}}
\medskip
\end{minipage}
\vspace{-1cm}
\caption{(Left) Evolution of 
$\hat{\mathbf{O}}$'s row support as a function of $\lambda_2$ -- black pixels 
along the $n$th row indicate that 
$\|\hat{\mathbf{o}}_n\|_2=0$, whereas white ones reflect that the responses 
from subject $n$ are deemed as outliers for given $\lambda_2$. The results for 
both the original and modified (introducing random and constant item 
responses) BFI datasets are shown.
(Right) Pseudo scree plot of outlier size
$(\|\hat{\mathbf{o}}_n\|_2)$; the 40 largest outliers are shown. Robust PCA 
declares the largest $8$ as aberrant responses.}
\label{fig:Fig_6} %\vspace{-1cm}
\end{figure}

Robust PCA is used to identify aberrant responses from real BFI data 
comprising the 
Eugene-Springfield community sample~\cite{Eugene_data}. The rows of 
$\mathbf{X}$ contain 
the $p=44$ item responses for each one 
of the $N=437$ subjects under study. For $q=5$, 
\eqref{eq:cost_l1unconstr} is 
solved over grid of $G_{\lambda}=200$ values of $\lambda_2$, where  
$\lambda_{\min}=10^{-2}\lambda_{\max}$, and $\lambda_{\max}=20$. 
The first plot of Fig. \ref{fig:Fig_6} (left) shows the evolution of 
$\hat{\mathbf{O}}$'s row support as a function of $\lambda_2$ with black 
pixels 
along the $n$th row indicating that 
$\|\hat{\mathbf{o}}_n\|_2=0$, and white ones reflecting that the responses 
from subject $n$ are deemed as outliers for the given $\lambda_2$. For example 
subjects $n=418$ and $204$ are strong outlier candidates due to random 
responding, since they enter the model 
($\|\hat{\mathbf{o}}_n\|_2>0$) for 
relatively large values of $\lambda_2$. The responses of e.g., subjects $n=63$ 
(all items rated `3') and $249$ ($41$ items rated `3' and $3$ items rated `4') 
are also undesirable, but are well modeled by \eqref{eq:factor_model} and are 
only deemed as outliers when $\lambda_2$ is quite small. These two 
observations 
are corroborated by the second plot of Fig. \ref{fig:Fig_6} (left), which 
shows the robust PCA results on a corrupted dataset, obtained from 
$\mathbf{X}$ by overwriting: (i) rows $151-160$ with random item 
responses drawn from a uniform distribution over $\{1,2,3,4,5\}$; and (ii) 
rows 
$301-310$ with constant item responses of value $3$.  

For $\lambda_2=5.6107$ corresponding to $\|\hat{\mathbf{O}}\|_0=100$, Fig. 
\ref{fig:Fig_6} (right) depicts the norm of the 40 largest outliers. 
Following the methodology outlined in Section \ref{ssec:synthetic}, 8 
subjects 
including $n=418$ and $204$ are declared as outliers by robust PCA. As a 
means of validating these results, the following procedure is adopted. Based 
on the BFI scoring key~\cite{BF_chapter}, a list of all pairs of items 
hypothesized to yield positively correlated responses is formed. For each $n$, 
one counts the `inconsistencies' defined as the number of times that subject 
$n$'s ratings for these pairs differ in more than four, in absolute value. 
Interestingly, after rank-ordering all subjects in terms of this inconsistency 
score, one finds that $n=418$ ranks highest with a count of $17$, $n=204$ 
ranks second ($10$), and overall the eight outliers found rank in the top 
twenty. 
  
\begin{figure}[h]
\begin{minipage}[b]{0.5\linewidth}
  \centering
  \centerline{\includegraphics[width=\linewidth]{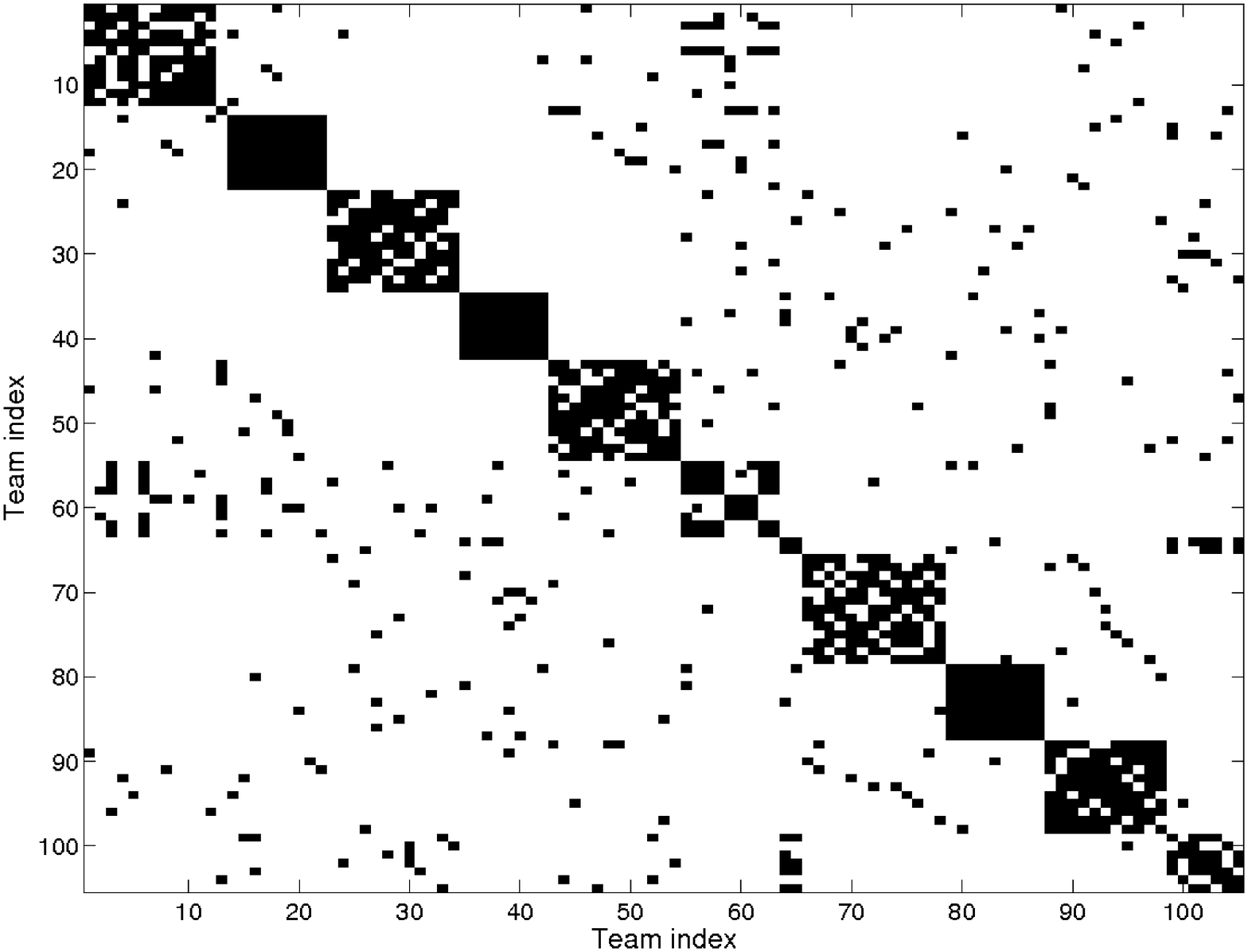}}
\medskip
\end{minipage}
\hfill
\begin{minipage}[b]{.5\linewidth}
  \centering
  \centerline{\includegraphics[width=\linewidth]{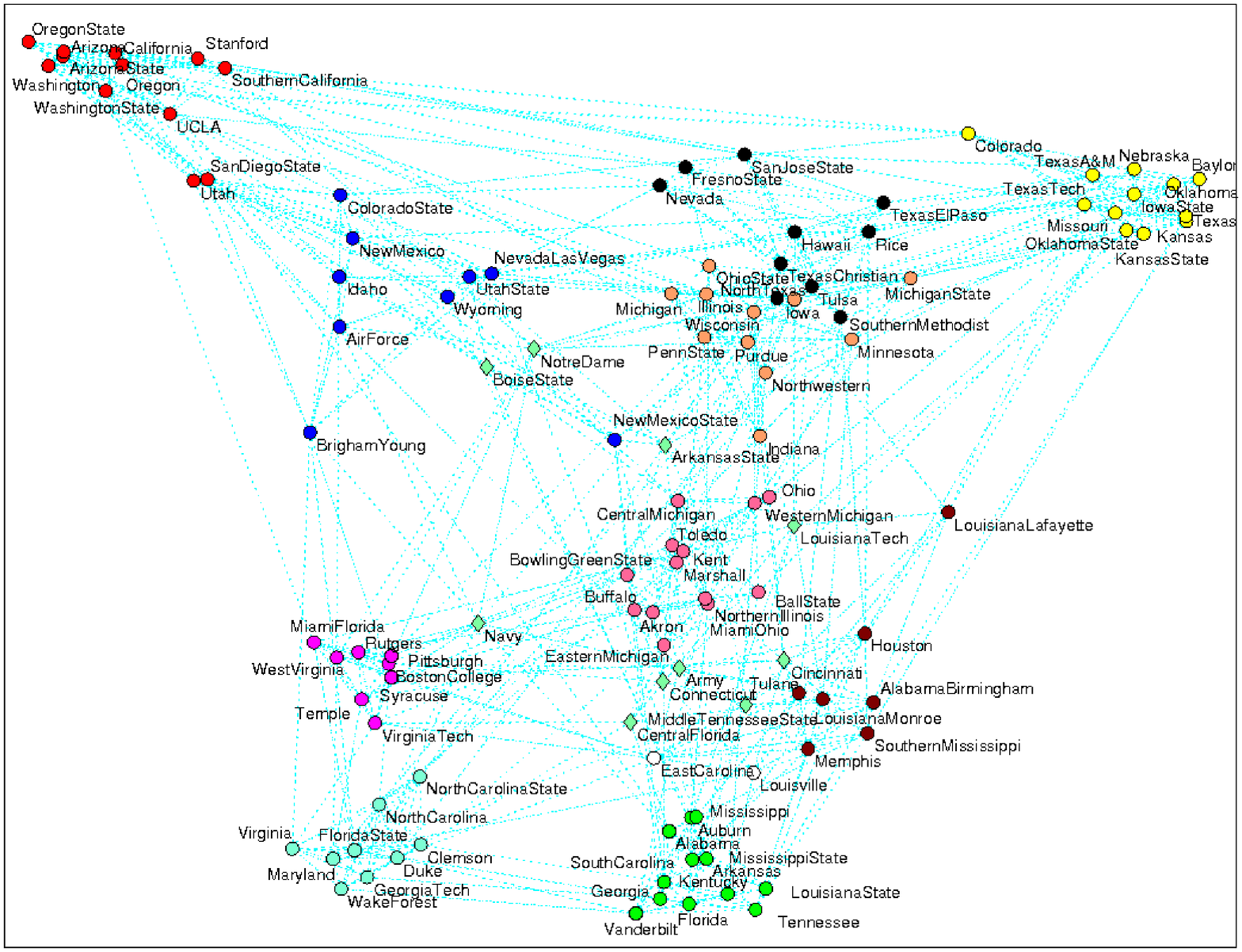}}
\medskip
\end{minipage}
%\vspace{-1cm}
\caption{(Left) Entries of $\mathbf{K}$ after removing the outliers, where 
rows and columns are permuted to reveal the clustering structure found by 
robust KPCA. 
(Right) Graph depiction  of the clustered network. Teams belonging to the same 
estimated 
conference (cluster) are colored identically. The outliers are represented 
as diamond-shaped nodes.}
\label{fig:Fig_5} %\vspace{-1cm}
\end{figure}

\noindent\textbf{Unveiling communities in social networks.} Next, robust KPCA 
is used to identify communities and outliers in a network of $N=115$ college 
football teams, by capitalizing on the connection between KPCA and spectral 
clustering~\cite[p. 548]{HTF09}. Nodes in the network graph represent teams 
belonging to eleven conferences (plus five independent teams), 
whereas (unweighted) edges joining pairs of nodes indicate that both teams 
played against each other during the Fall 2000 Division I 
season~\cite{network_data}. The kernel matrix used to run robust KPCA is 
$\mathbf{K}=\zeta\mathbf{I}_N+\mathbf{D}^{-1/2}\mathbf{A}\mathbf{D}^{-1/2}$, 
where $\mathbf{A}$ 
and $\mathbf{D}$ denote the graph adjacency and degree matrices, respectively; 
while $\zeta>0$ is chosen to render $\mathbf{K}$ positive semi-definite. The 
tuning parameters are chosen as $\lambda_2=1.297$ so that
$\|\hat{\mathbf{O}}\|_0=10$, while $\lambda_\ast=1$, and $\bar{q}=3$. Fig. 
\ref{fig:Fig_5} (left) shows the entries of $\mathbf{K}$, where rows and 
columns are permuted to reveal the clustering structure found by robust KPCA 
(after removing the outliers); see also Fig. \ref{fig:Fig_5} (right). The 
quality of the clustering is assessed through the adjusted rand index (ARI) 
after excluding outliers~\cite{pf_vk_gg_clustering}, which yielded the value 
0.8967. Four of the teams deemed as outliers are Connecticut, Central Florida, 
Navy, and Notre Dame, which are indeed teams not belonging to any major 
conference. The community structure of traditional powerhouse conferences such 
as Big Ten, Big 12, ACC, Big East, and SEC was identified exactly.

% % % % % % % % % % % % % % % % % % % % % % % % % % % % % % % % % % % % % % % %
%                         Section VIII                                        %
% % % % % % % % % % % % % % % % % % % % % % % % % % % % % % % % % % % % % % % %

\section{Concluding Summary}\label{sec:conclusion}

Outlier-robust PCA methods were developed 
in this paper, to obtain low-dimensional representations of 
(corrupted) data. Bringing together the seemingly unrelated 
fields of robust statistics and sparse regression, the novel robust PCA 
framework was found rooted at the crossroads of outlier-resilient estimation, 
learning via (group-) Lasso and kernel methods, and real-time 
adaptive signal processing. Social network 
analysis, video surveillance, and psychometrics, were highlighted as relevant 
application domains.\\

\noindent\textbf{Acknowledgment:} The authors would like to thank Prof. Niels 
Waller 
(Department of Psychology, University of Minnesota) for the fruitful 
discussions on IRT and the measurement of the Big Five; and Dr. Lewis Goldberg
(Oregon Research Institute) for facilitating access to the BFI data studied in 
Section \ref{ssec:real}.

%------------------------------------------------------------------------------
%----------------------------TheAppendix---------------------------------------
%%-----------------------------------------------------------------------------
{\Large\appendix}

Towards establishing the equivalence between problems 
\eqref{eq:cost_l1unconstr}
and \eqref{eq:variational_w_rho},
consider the pair $\{\hat{\mathcal{V}},\hat{\mathbf{O}}\}$ that solves
\eqref{eq:cost_l1unconstr}. Assume that $\hat{\mathcal{V}}$ is given, and
the goal is to determine $\hat{\mathbf{O}}$.
Upon defining the residuals $\hat{\mathbf{r}}_n:=\mathbf{x}_n-\hat{\mathbf{m}}-\hat{\mathbf{U}}
\hat{\mathbf{s}}_n$ and from the row-wise decomposability of
$\|\cdot\|_{2,r}$, the rows
of $\hat{\mathbf{O}}$ are separately given by
\begin{equation}\label{eq:cost_oi_convex}
\hat{\mathbf{o}}_n:=\arg\min_{\mathbf{o}_n\in\mathbb{R}^p}
\left[\|\hat{\mathbf{r}}_n-\mathbf{o}_n\|_2^2+\lambda_2\|\mathbf{o}_n\|_2\right],\quad
n=1,\ldots,N.
\end{equation}
For each $n=1,\ldots,N$, because \eqref{eq:cost_oi_convex}
is nondifferentiable at the origin one should consider two
cases: i) if $\hat{\mathbf{o}}_n=\mathbf{0}_p$, it follows that the minimum cost in
\eqref{eq:cost_oi_convex} is $\|\hat{\mathbf{r}}_n\|_2^2$; otherwise, ii) if
$\|\hat{\mathbf{o}}_n\|_2> 0$, the first-order condition for optimality
gives $\hat{\mathbf{o}}_n=\hat{\mathbf{r}}_n-
(\lambda_2/2)\hat{\mathbf{r}}_n/\|\hat{\mathbf{r}}_n\|_2$ provided 
$\|\hat{\mathbf{r}}_n\|_2>\lambda_2/2$, and the minimum cost is 
$\lambda_2\|\hat{\mathbf{r}}_n\|_2-\lambda_2^2/4$. Compactly, the solution 
of \eqref{eq:cost_oi_convex} is given by $\hat{\mathbf{o}}_n=
\hat{\mathbf{r}}_{n}(\|\hat{\mathbf{r}}_{n}\|_2-\lambda_2/2)_+/\|\hat{\mathbf{r}}_{n}\|_2$
, while the minimum cost in \eqref{eq:cost_oi_convex} after minimizing w.r.t. 
$\mathbf{o}_n$ is $\rho_v(\hat{\mathbf{r}}_n)$ [cf. \eqref{eq:rho_def} and
the argument following \eqref{eq:cost_oi_convex}].
The conclusion is that $\hat{\mathcal{V}}$
is the minimizer of 
\eqref{eq:variational_w_rho}, in addition to being the solution
of \eqref{eq:cost_l1unconstr} by definition.

%---------------------------------The References--------------------------------------------------------------------------
\newpage
\bibliographystyle{IEEEtranS}
\bibliography{IEEEabrv,biblio}

%---------------------------------The Figures-----------------------------------------------------------------------------
%\vspace{-0.5cm}
 \begin{table}[h]
 \renewcommand{\arraystretch}{1.3}
%\caption{Results for the first synthetic data test}
 \caption{}
 \label{table:results} \centering
 \begin{tabular}{|c|c|c|c|c|}
 \hline
 \bfseries $\sigma_e^2$ & \bfseries $\lambda_2^{\ast}$ in 
\eqref{eq:cost_l1unconstr}& 
 \bfseries $\bar{\textrm{err}}$ for
 \eqref{eq:cost_l1unconstr} (refined)& \bfseries $\bar{\textrm{err}}$ for
\eqref{eq:spcp}& \bfseries $\bar{\textrm{err}}$ for
PCA\\
\hline\hline
$0.01$ & $0.7142$ & $0.0622$ & $0.0682$ & $0.4679$ \\\hline
$0.05$ & $1.7207$ & $0.1288$ & $0.1519$ & $1.0122$ \\\hline
$0.1$ & $2.4348$ & $0.1742$ & $0.2150$ & $1.4141$ \\\hline
$0.25$ & $3.6084$ & $0.2525$ & $0.3403$ & $2.2480$ \\\hline
$0.5$ & $6.1442$ & $0.3361$ & $0.4783$ & $3.1601$ \\\hline
 \end{tabular}
 \end{table}

%\vspace{-1cm}
%\newpage
%
%

%\vspace{-4cm}

\begin{figure}[h]
\begin{center}
\includegraphics[width=0.8\linewidth]{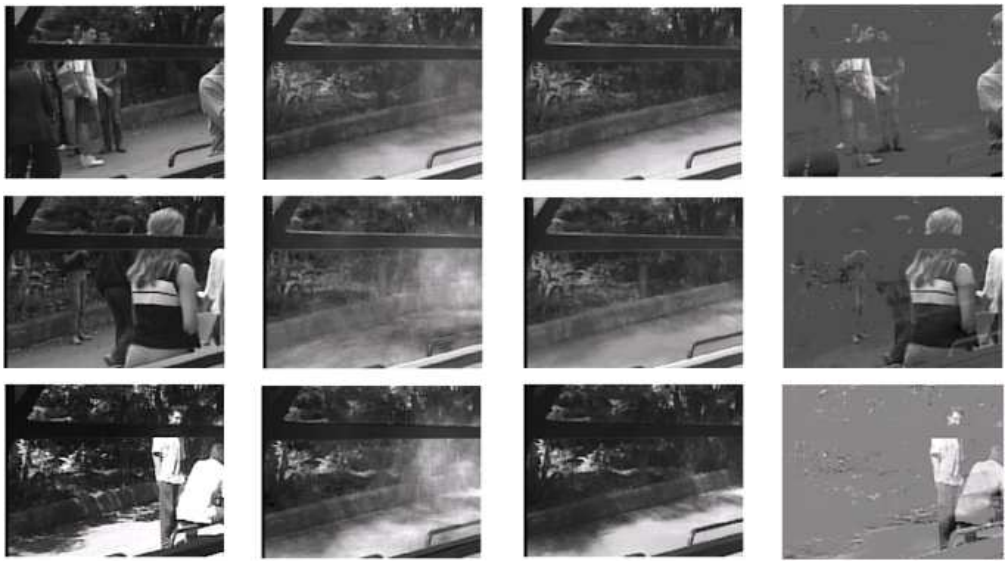}\\
%\vspace{-0.5cm}
\caption{Background modeling for video
surveillance. First column: original frames. Second column: PCA
reconstructions, where the presence of undesirable `ghostly' artifacts is apparent, since
PCA is not able to completely separate the people from the
background. Third column: robust PCA
reconstructions, which recover the illumination changes while
successfully subtracting the people. Fourth column: outliers in
$\hat{\mathbf{o}}$, which mostly
capture the people and abrupt changes in illumination.}\label{fig:Fig_4}
\end{center}
\end{figure}

\end{document}